\PassOptionsToPackage{dvipsnames}{xcolor}
\documentclass{article} % For LaTeX2e
\usepackage[final]{colm2025_conference}

\usepackage{microtype}
\usepackage{hyperref}
\usepackage{url}
\usepackage{booktabs}

\usepackage{lineno}

\usepackage{natbib}

\usepackage{hyperref}

\usepackage[disable]{todonotes}
\usepackage{amsmath}
\usepackage{amssymb}
\usepackage{bm}

\usepackage{array}
\usepackage{booktabs}
\usepackage{multirow}
\usepackage{hhline}

\usepackage{caption}
\usepackage{subcaption}

\usepackage{makecell}

\usepackage[inline]{enumitem}

\usepackage{wrapfig}

\usepackage{placeins}

\usepackage{graphicx} % for rotatebox

\usepackage{listings}

\usepackage{ulem}

\usepackage{pgf}

\usepackage{colortbl} % <-- also imports xcolor
\usepackage{xstring}

\usepackage{stringstrings}

\usepackage{collcell}

\usepackage{latexml}
\let\vec\mathbf
\newcommand*{\PLM}{P_\lambda}

\makeatletter
\newcommand{\Xh}[1][\@empty]{%
  In\ New\ York,\ days\ are\ % ends in space
  \ifx#1\@empty
     [not]%
  \else%
     \ifx#1-%
        \textbf{not}%
     \fi
  \fi
  \ shortest\ in\ December%
}
\makeatother

\makeatletter
\newcommand{\Xp}[1][\@empty]{%
  December\ is\ %                ← single intended space
  \ifx#1\@empty
     [not]%
  \else%                         ← % kills the unwanted line-break space
     \ifx#1-%
        \textbf{not}%
     \fi
  \fi
  \ during\ the\ winter\ for\ New\ York%
}
\makeatother

\newcolumntype{A}{ >{$} r <{$} @{} >{${}} l <{$} } % A for "align"

\newcommand*{\MinNumber}{0}
\newcommand*{\MidNumber}{0.5}
\newcommand*{\MaxNumber}{1}

\definecolor{lightgray2}{RGB}{220,220,220}

\newcommand*{\MinColor}{BurntOrange}
\newcommand*{\MidColor}{lightgray2}
\newcommand*{\MaxColor}{Cyan}

\newcommand{\ApplyGradient}[1]{%
    \getargs[q]{#1}%
    \ifdim \narg pt>0 pt%%%% if cell is empty don't do anything.
        \edef\argnumber{\argi}
        \IfDecimal{\argnumber}{% if first argument is something other than a number, just output it.
            \ifdim \argnumber pt>\MidNumber pt%
                \pgfmathsetmacro{\PercentColor}{max(min(100.0*(\argnumber-\MidNumber)/(\MaxNumber-\MidNumber),100.0),0.00)}%
                \ifdim \narg pt<2 pt%% no override
                    \edef\x{\noexpand\cellcolor{\MaxColor!\PercentColor!\MidColor}}%
                \else%% color override
                    \edef\x{\noexpand\cellcolor{\argii}}%
                \fi%
                \x\argnumber%
            \else%
                \pgfmathsetmacro{\PercentColor}{max(min(100.0*(\MidNumber-\argnumber)/(\MidNumber-\MinNumber),100.0),0.00)}%
                \ifdim \narg pt<2 pt%% no override
                    \edef\x{\noexpand\cellcolor{\MinColor!\PercentColor!\MidColor}}%
                \else%% color override
                    \edef\x{\noexpand\cellcolor{\argii}}%
                \fi%
                \x\argnumber%
            \fi%
        }{\argnumber}%
    \else%
    \fi%
}

\iflatexml
\newcolumntype{R}{c}
\else
\newcolumntype{R}{>{\collectcell\ApplyGradient}c<{\endcollectcell}}
\fi

\makeatletter
\pretocmd{\appendix}{%
}{}{}
\makeatother

\makeatletter
\AtBeginDocument{%
  \begingroup
  \normalsize
  \let\tmp@n@s\f@size
  \let\tmp@n@b\f@baselineskip
  \small
  \let\tmp@s@s\f@size
  \let\tmp@s@b\f@baselineskip
  \xdef\semismall@size{\fpeval{(\tmp@n@s+\tmp@s@s)/2}}%
  \xdef\semismall@baselineskip{\fpeval{(\tmp@n@b+\tmp@s@b)/2}}%
  \endgroup
}
\newcommand{\semismall}{\fontsize{\semismall@size}{\semismall@baselineskip}\selectfont}

\makeatletter
\AtBeginDocument{%
  \begingroup
  \footnotesize
  \let\tmp@n@s\f@size
  \let\tmp@n@b\f@baselineskip
  \scriptsize
  \let\tmp@s@s\f@size
  \let\tmp@s@b\f@baselineskip
  \xdef\semiscript@size{\fpeval{(\tmp@n@s+\tmp@s@s)/2}}%
  \xdef\semiscript@baselineskip{\fpeval{(\tmp@n@b+\tmp@s@b)/2}}%
  \endgroup
}
\newcommand{\semiscript}{\fontsize{\semiscript@size}{\semiscript@baselineskip}\selectfont}

\usepackage{amsmath,amsfonts,bm}

\def\eqref#1{equation~\ref{#1}}
\def\1{\bm{1}}

\DeclareMathAlphabet{\mathsfit}{\encodingdefault}{\sfdefault}{m}{sl}
\SetMathAlphabet{\mathsfit}{bold}{\encodingdefault}{\sfdefault}{bx}{n}

\newcommand{\E}{\mathbb{E}}

\DeclareMathOperator*{\argmin}{arg\,min}

\makeatletter
\renewcommand{\@makefnmark}{\raisebox{-0.2ex}{\textsuperscript{\@thefnmark}}}
\makeatother

\definecolor{darkblue}{rgb}{0, 0, 0.5}
\hypersetup{colorlinks=true, citecolor=darkblue, linkcolor=darkblue, urlcolor=darkblue}

\title{Truth-value judgment in language models:\\`truth directions' are context sensitive}

\author{Stefan F. Schouten, Peter Bloem, Ilia Markov, Piek Vossen \\
Vrije Universiteit Amsterdam\\
\texttt{\{s.f.schouten,p.bloem,i.markov,p.t.j.m.vossen\}@vu.nl} \\
}

\begin{document}

\ifcolmsubmission
\linenumbers
\fi

\maketitle

\begin{abstract}
% general motivation
Recent work has demonstrated that the latent spaces of large language models (LLMs) contain directions predictive of the truth of sentences.
% problem formulation
Multiple methods recover such directions and build probes that are described as uncovering a model's ``knowledge'' or ``beliefs''.
% proposal
We investigate this phenomenon, looking closely at the impact of \textit{context} on the probes.
Our experiments establish where in the LLM the probe's predictions are (most) sensitive to the presence of related sentences, and how to best characterize this kind of sensitivity.
% method
We do so by measuring different types of consistency errors that occur after probing an LLM whose inputs consist of hypotheses preceded by (negated) supporting and contradicting sentences.
We also perform a causal intervention experiment, investigating whether moving the representation of a premise along these \textit{truth-value directions} influences the position of an entailed or contradicted sentence along that same direction.
% results
We find that the probes we test are generally context sensitive, but that contexts which should not affect the truth often still impact the probe outputs.
% Our experiments suggest that in earlier layers the context is assumed to be true, whereas in later layers the LLM's degree of belief in the premise plays a larger role.
Our experiments show that the type of errors depend on the layer, the model, and the kind of data.
Finally, our results suggest that truth-value directions are causal mediators in the inference process that incorporates in-context information.

\end{abstract}

\section{Introduction} \label{sec:introduction}
%
% P1  -  identify motivating problem using example
%
As Large Language Models (LLMs) enjoy increasing mainstream adoption, it becomes more important to understand why they fail in some cases, while excelling in others. 
Recent findings show that LLM latent spaces contain directions predictive of the truth of sentences \citep{burns_discovering_2023, marks_geometry_2024}. 
Probes that leverage these directions to assign truth values to sentences are accurate even in misleading contexts where prompting fails.
% Given a sentence, the activations for its last token can be projected onto such a direction to assign the sentence a truth value.
% Recent findings establish that probes can be trained to assign truth-values to sentences based on LLM activations.
% Such \textit{truth-value probes} can work even in misleading contexts where prompting fails \citep{burns_discovering_2023}.
When considering simple declarative sentences on their own, it makes sense to evaluate such \textit{truth-value probes} primarily by their accuracy. 
% The correct truth-value of a simple declarative sentence is typically easily determined. 
For example, we might have a hypothesis: \textit{``\Xh[+]''}, which we would expect an LLM to represent as true.
But how should we evaluate if this hypothesis is placed in the context of an incorrect premise, like: \textit{``\Xp[-]''}?
In that case, we no longer necessarily have the same expectations.
For example, if the premise is understood as setting up a counterfactual scenario, then we would expect an LLM to represent the hypothesis as false.
% We now expect that either: 
% (1) the premise is disbelieved and the hypothesis is evaluated a priori (as though the premise was not there); or
% (2) the hypothesis is evaluated counterfactually.
In other words, for multi-sentence inputs, evaluations must prioritize \textit{coherence}: the degree of logical consistency in truth-value assignments.

%
% P2  -  explain importance of solving the problem
%
% When an LLM's activations for a sentence $S$ appear in the ‘true part’ of its latent space, it is tempting to refer to it as \textit{believing} $S$. 
When a truth-value probe assigns a sentence $S$ the label \textsc{true}, it is tempting to refer to the model as \textit{believing} $S$. 
Especially because such explanations of LLM behavior are highly plausible \citep[convincing to humans, ][]{jacovi_towards_2020}. 
Explaining LLMs by appealing to attitudes like belief has recently been endorsed as \textit{propositional interpretability} \citep{chalmers_propositional_2025}, and truth-value probes seem like a promising method. Specifically, they could serve to gauge if an LLM---given one or more sentences as input---{judges} them to be true (actively believes them) or not.
But, if a probe's truth-value assignments are not coherent, it has failed to reveal (representations of) judgment or belief \citep{herrmann_standards_2025}, meaning an interpretation that appeals to those concepts would be unfaithful.

%
% P3  -  outline how we approach the problem
%
In this work, we study coherence by probing LLMs on inputs where hypotheses are preceded by  premises which appear either affirmed or negated. 
We take inspiration from the \textit{truth-value judgment} task in language acquisition research where subjects are “asked to make a bipolar judgment about whether a statement accurately describes a particular situation alluded to in some context or preamble” \citep{mcdaniel_truth-value_1996}.
We find that {truth-value assignments} are context sensitive, but also sensitive to irrelevant information.

We also investigate if the relevant directions in latent-space causally mediate truth-value judgment, or if they only reflect the outcome of that process.
Specifically, we establish if a representation's position along a truth-value direction determines (in part) where subsequent statements are positioned along the same direction. 
% \todo{expand upon with latest results}The different probing methods show similar levels of overall consistency.
% Our experiments further show that the type of made does depend on the layer, the (type of) model, and the kind of data. 
Our results suggest that these directions are mediators in the inference process that incorporates in-context information.

In summary, our contributions are: 
(1) experiments evaluating the context sensitivity of {truth-value probes} and the consistency with which they incorporate it; we quantify across layers, model sizes (7 and 13 billion), and type of training (pretrained-only vs. instruction-tuned); and
(2) an experiment demonstrating that {truth-value directions} causally mediate natural language inference.
We also propose a new variant of CCS \citep{burns_discovering_2023} for which convergence is more stable, and otherwise behaves and performs similarly. 
Our code is available at
% \url{https://anonymous.4open.science/r/lcb2-AF5D}%
\url{https://github.com/sfschouten/tvj-in-llms}%
.

\begin{figure}[t]
    \centering
    % \vspace{-4mm}
    \includegraphics[width=\linewidth]{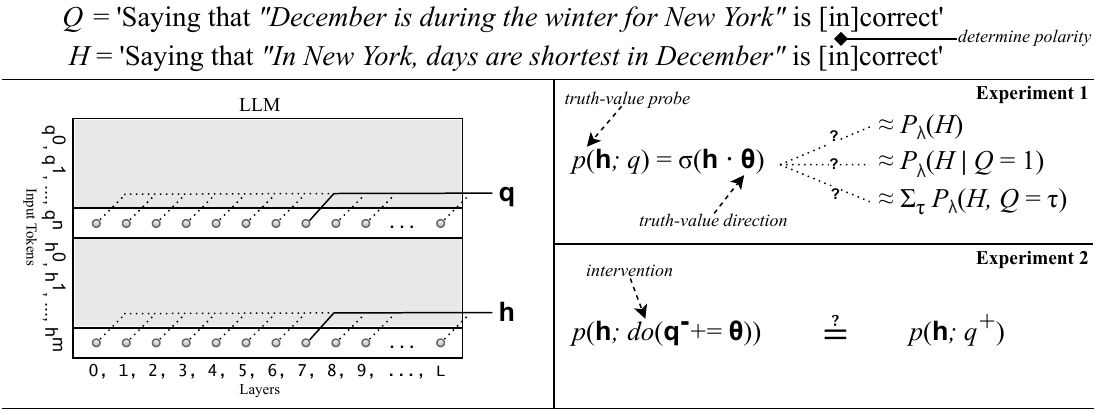}
    \vspace{-6mm}
    \caption{Overview of our setup. LLM representations $\vec{q}$ and $\vec{h}$ for a premise and hypothesis are extracted and used to train truth-value probes. In experiment 1, the probes are evaluated to determine if and how they incorporate context. In experiment 2, we move a premise's representation in the identified truth-value direction, measuring if the probability assigned to the hypothesis changes accordingly.}
    \label{fig:setting_overview}
\end{figure}
\section{Related Work}
% intro
Probing LLM representations for the truth of sentences has recently received much interest.
\cite{burns_discovering_2023} introduce Contrast Consistent Search (CCS), an unsupervised probing methods based on the representations of contrasting sentence pairs. 
Their probes often outperform a (zero-shot) prompting approach, even when applied to misleading prompts. 
% One of these methods (Contrast Consistent Search, CCS), when used with a linear probe, finds a direction in latent space that is optimized such that probabilities assigned to the contrasting sentences sum up to one. 
%
\cite{li_inference-time_2023} shift model activations in the `truth direction' at inference time, mitigating hallucination. 
Their interventions use 
1) directions from probes trained with logistic regression (LR) and CCS; and 
2) a new method (Mass Mean Shift), which finds the direction as the difference between the means of the true and false sentence activations. 
%
% \cite{azaria_internal_2023} train truthfulness probes on a new dataset of out-of-distribution text constructed from tabular data. 
% They find that probes trained for truthfulness also correlate with sentence length and the frequencies of words. 
% But, their probe nonetheless outperforms the language model's predictions.
%
\cite{marks_geometry_2024} take Mass Mean Shift and use its directions to build probes (Mass-Mean Probing, MMP). 
% \cite{marks_geometry_2023} introduce Mass-Mean Probing (MMP), based on Mass Mean Shift method by \citet{li_inference-time_2023}. 
% This method finds the belief direction by taking the difference between the means of the true and false sentence representations. 
% They introduce new datasets, some of which curated to include only statements that are ``uncontroversial, unambiguous, and simple enough that our LLMs are likely to understand them''.
They show all probes (based on LR, CCS, and MMP) generalize well between datasets, with MMP performing the best. 
In a causal intervention experiment, they move representations in the identified directions, showing that MMP is the best mediator: it increases the probability of the model calling a false statement true the most.

Previous work has used both data consisting of single facts and longer inputs, such as various NLI datasets, but did not study the impact of the context. 
We specifically study the in-context behavior of truth-value probes, analyzing their consistency, and what this means for the way LLMs incorporate contextual information.
Like \citet{marks_geometry_2024}, we also investigate the causal implication of directions in LLM latent space.
However, rather than investigate what causes the greatest change in token predictions, we investigate which direction to move a premise in, such that it causes the correct change in the probability of a related hypothesis, as evaluated by the same direction.

% \citet{herrmann_standards_2025} have recently formulated four requirements for a representation to count as belief-like.
% One of those requirements is \textit{coherence}, which requires that belief probes be logically consistent.
% Our method measures specific kinds of coherence: 
% two error scores measure the extent to which probed beliefs depend on semantically irrelevant factors, and the two others measure if beliefs are consistent with semantically relevant contexts in one of two ways. 

Recent work has also criticized this type of probing. 
% \citet{levinstein_still_2023} reproduce the work of \cite{burns_discovering_2023} and \citet{azaria_internal_2023}, and show their probes do not generalize to negated variants of the data, even when that negated data is also present during training.
% On this matter the results of \citet{levinstein_still_2023} and \citet{marks_geometry_2023} seem contradictory.
% \citeauthor{levinstein_still_2023} further point out conceptual problems, regarding the difficulty of isolating truth as a feature, especially using only coherence properties (such as probabilities summing up to one).
Truth-value probes might identify properties that correlate with truth \citep{levinstein_still_2024}, especially when truth is not the most salient feature \citep{farquhar_challenges_2023}.
%such as commonly held beliefs 
% Further criticism was provided by \citet{farquhar_challenges_2023}, specifically regarding unsupervised approaches such as CCS.
% Their analysis includes a theoretical result showing that for any CCS probe there is an arbitrary probe with identical loss.
% They also report empirical results showing that depending on the prompt, or text or the placement of the probe in the text, 
% This is also supported by \cite{farquhar_challenges_2023}, whose experiments demonstrate that unsupervised probes fail to reveal ``knowledge'' when truth is not the most salient feature. 
%
% A common theme in these criticisms is the possibility of the truth of sentences being confounded with other features.
% Specifically, both point out that these probing methods do not automatically find an objective evaluation of truth by the LLM, but could also find evaluations of the truth from another perspective. For example, how commonly held the belief in the sentence is, or what a person mentioned in the text would believe about the sentence.
%
% We think it's important to distinguish between arbitrary spurious correlations, and beliefs that cannot be said to be the LLM's own. % (e.g. commonly-held beliefs, or beliefs belonging to people or characters mentioned in the text as in \citet{farquhar_challenges_2023} or \citet{zhu_language_2024}). 
% We argue our method is resistant against both. 
We evaluate the probes' coherence, which we do not expect spurious correlations to exhibit.

\section{Methodology}
We describe our method in three parts:
in \ref{ssec:belief_probing} we cover truth-value probes, the necessary assumptions, notation, and methods to construct them;
in \ref{ssec:truth_value_judgment} we describe how to construct samples and the possible quantities measured by probes in our setting; and
in \ref{ssec:evaluation} we introduce error scores with which we measure (different kinds of) coherence. 

\subsection{Truth-value probes}\label{ssec:belief_probing}
We use several probing methods in our experiments.
These methods use datasets of sentences, consisting of both true and false statements. 
We can turn any true statement into a false statement (and vice versa) by negating it. 
% We denote the affirmative case with $X^+$ and the negative case as $X^-$, and their LLM vector representations are given as $\vec{x}^+$, $\vec{x}^-$ 
% \underline{\smash{We use $X^+$ and $X^-$ to denote the affirmed (positive) and negated (negative) case as $X^-$}}, and their LLM vector representations are given as $\vec{x}^+$, $\vec{x}^-$ 
{We use superscript $+$ and $-$ to denote the affirmed ($X^+$) and negated ($X^-$) variants of a sentence, respectively.} 
Their LLM vector representations are given in bold, i.e. $\vec{x}^+$, $\vec{x}^-$
(see \autoref{ssec:experiments:data} for how we negate sentences and for how vector representations are extracted).
Thus, the dataset used to train probes consist of pairs of hidden states extracted for the positive and negative variants of statements $(\vec{x}^+, \vec{x}^-, y^+, y^-) \in \mathcal{D}$, and their labels indicating which of the two is true (with $y^+ = 1 - y^-$).
% When we refer to $X$ or $\vec{x}$ without polarity, the polarity could be positive or negative.
% \underline{\smash{Note that statements can be true or false regardless of their polarity}}, i.e. whether they are affirmed or negated.

When using truth-value probes, we assume that the truth of sentences is latently modeled by LLMs. 
We characterize this latent ($\lambda$) model as a probability distribution $\PLM(X)$.
%\footnote{This distribution is entirely separate from the probabilities assigned to tokens by the LM-head.}
The probes $p(\vec{x})$ are assumed to (approximately) recover this distribution.
We believe these assumptions are fair, because previous work \citep{burns_discovering_2023,marks_geometry_2024} has found directions in latent spaces that suggest LLMs do track sentence truth.

%: $\PLM(X) \sim \textup{Bernoulli}(p'), \text{ where } p' \approx p(\vec{x})$. 
% We use $\PLM(x)$ and $\PLM(\neg x)$ as short for $\PLM(X=1)$ and $\PLM(X=0)$.
% We use $\PLM(x)$ as a shorthand for $\PLM(X{=}1)$.
%
Probes are constructed as: $p(\vec{x}) = \sigma(\vec{x} \cdot \bm{\theta})$, where $\bm{\theta}$ is the truth-value direction and $\sigma$ is the sigmoid. 
Because the probes do not have bias terms, all inputs are mean-normalized (in line with previous work):
% \begin{equation}
$
    \vec{x}^+ \mathrel{-}= \bm{\mu} \text{ and }
    \vec{x}^- \mathrel{-}= \bm{\mu}, \text{ where } \bm{\mu} 
  = \frac{1}{2|\mathcal{D}|}\sum_{(\mathbf{x}^+, \mathbf{x}^-) \in \mathcal{D}} \;( \mathbf{x}^+ + \mathbf{x}^- )
$.
% \end{equation}

\paragraph{Mass Mean Probing (MMP)}
is a supervised method, which defines the truth-value direction as the difference between the average of the correct and incorrect statements:
\begin{align}
    \bm{\theta_{\text{mm}}} &= 
    \E_{\vec{x}, y} [ \vec{x} \,|\, y=1 ] - \E_{\vec{x}, y} [ \vec{x} \,|\, y=0],
\end{align}
where $y$ is the truth-value (label) for the statement $X$.
% Probes are then constructed as 
%follows 
% $
% % \begin{align}
%     p(\vec{x}) = \sigma(\bm{\theta_{\text{mm}}}^\intercal \vec{x})
% % \end{align}
% $.
We do not include the version of MMP that requires an i.i.d. assumption, because we also evaluate on out-of-distribution data.

\paragraph{Logistic Regression (LR)} is also used to train a supervised probe. 
They are trained on $\vec{x}' = \vec{x}^- - \vec{x}^+$, i.e. the difference between the negative and positive statements.
% We use LR without a bias/intercept term:
{\setlength{\belowdisplayskip}{0pt} 
\begin{align}
    \bm{\theta_{\text{lr}}} = \argmin_{\bm{\theta}} - \E_{\vec{x}', y^+} \left[ 
        y^+ \ln \sigma(\bm{\theta} \cdot \vec{x'}) 
  + (1-y^+) \ln\left( 1 - \sigma(\bm{\theta} \cdot \vec{x'}) \right)
    \right],
\end{align}}
where $y^+$ is the label for the positive variant of the sample, i.e. whether $X^+$ is true.
% For all other hyperparameters we use the default values of SKLearn \citep{pedregosa_scikit-learn_2011}. 
%This method learns a hyperplane that separates true facts from false facts, and its normal is used as the belief direction.

\paragraph{Contrast Consistent Search (CCS)}
is an unsupervised\footnote{By unsupervised we mean that no knowledge of which sentences are true or false is given.} method 
%where the probe is optimized such that:
%the probabilities assigned to $x^+$ and $x^-$ sum up to one:
% \begin{align}
% $
    % p(\vec{x}^+)\, + \,p(\vec{x}^-) = 1. 
% $
    % \label{eq:ccs_main}
% \end{align}
%
% There is a degenerate solution when 
% . \label{eq:ccs_degen}
% Thus, the objective has two terms:
with as its objective:
{\setlength{\belowdisplayskip}{0pt} \setlength{\abovedisplayskip}{3pt} 
\begin{align}
    \bm{\theta_{\text{ccs}}} \; = \;
    \argmin_{\bm{\theta}} \;
    \E_{\vec{x}^+, \vec{x}^-}
        \left[\,[ 1 - p(\vec{x}^+) - p(\vec{x}^-) ]^2  
    \; + \;
        \min\{ p(\vec{x}^+),\,p(\vec{x}^-) \}^2 \,
    \right],
    \label{eq:ccs_objective}
\end{align}}
which has two terms: the consistency-loss (encouraging solutions where the probabilities add up to one), and the confidence-loss (encouraging non-degenerate solutions, i.e. $ p(\vec{x}^+) \neq p(\vec{x}^-) \neq 0.5 $).
% The first term 
The objective can be understood as finding a hyperplane with normal $\bm{\theta}$ that, for each pair: (1) separates $\vec{x}^+$ from $\vec{x}^-$, and (2) is equidistant to $\vec{x}^+$ and $\vec{x}^-$.
% The belief direction found by this method is this normal $\bm{\theta_\text{ccs}} = \vec{w}$. %(also see \autoref{apx:ccs_interpretation}).

\paragraph{Contrast Consistent Reflection (CCR)} 
is proposed here as a variant of CCS. 
Rather than finding a hyperplane from which $\vec{x}^+$ and $\vec{x}^-$ are equidistant, this method requires $\vec{x}^+$ and $\vec{x}^-$ to be each other's reflection in the hyperplane.
It has the following objective: 
{\setlength{\belowdisplayskip}{0pt} 
\begin{align}
    \bm{\theta_{\text{ccr}}} \; = \;
    \argmin_{\hat{\bm{\theta}}} \;
    \E_{\vec{x}^+, \vec{x}^-} \left[ \;
        ||\vec{x}^+ -  \vec{P}\vec{x}^- ||_2
    \; \right], 
    % \text{where }
    % \vec{P} = \vec{I} - 2\vec{\bm{\theta}}\vec{\bm{\theta}}^\intercal
    \label{eq:ccr_objective}
\end{align}}
% where $\hat{\bm{\theta}}$ is a unit vector and $\vec{P} = \vec{I} - 2\hat{\bm{\theta}}\hat{\bm{\theta}}^\intercal$ is the Householder transformation performing the reflection.
where unit vector $\hat{\bm{\theta}}$ determines the direction of the Householder reflection $\vec{P} = \vec{I} - 2\hat{\bm{\theta}}\hat{\bm{\theta}}^\intercal$.
This objective does not share the degenerate solution of CCS. 
This is because for $p(\vec{x}^+) = p(\vec{x}^-) = 0.5$, we need $\bm{\theta} \cdot \vec{x}^+ = \bm{\theta}\cdot \vec{x}^- = 0$, and since $|\bm{\theta}| = 1$ this would imply that $\bm{\theta}$ is orthogonal to $\vec{x}^+$ and $\vec{x}^-$. 
Thus, while they are equidistant in that scenario (a distance of zero), assuming that $\vec{x}^+ \neq \vec{x}^-$, they will not be each other's reflection. 

On our data CCS does not consistently converge to a good minimum (see \autoref{apx:ccr_vs_ccs}).
% We follow \citet{burns_discovering_2023} in training 10 probes and using the probe with the lowest training loss.
% However, we find that this procedure nonetheless produces directions that vary considerably from layer to layer (see \autoref{apx:cosines}) making it harder to analyse.
CCS finds directions that vary considerably from layer to layer (see \autoref{apx:cosines}) making it harder to analyze.
% CCR's objective has one term, and we have found it to achieve similar performance with more stable convergence, without the need to train multiple probes. 
We see CCR achieve similar performance with more stable convergence. %, without the need to train multiple probes. 
By including CCR, we can see how unsupervised methods compare to supervised methods, without having to worry about observations being artifacts of CCS's instability.

% \begin{table}[t]
%     \newcolumntype{s}{>{\small}c}
%     \newcolumntype{A}{>{$}c<{$}}
%     \centering
%     \begin{tabular}{Ass}
%     \toprule
%        & \textsc{base statement} & \textsc{meta statement} \\
%     \midrule
%      Q & \Xp{}.   & Saying that ``\Xp[+]{}'' is [in]correct. \\
%      H & \Xh{}.   & Saying that ``\Xh[+]{}'' is [in]correct. \\
%     \bottomrule
%     \end{tabular}
%     \caption{An example premise-hypothesis pair, showing how we wrap sentences in meta statements.}
%     \label{tab:example1}
% \end{table}

\subsection{Probing for truth-value judgment}\label{ssec:truth_value_judgment}
% Truth-value judgment (TVJ) tasks are used in language acquisition research to assess children's linguistic competencies. 
% %
% Subjects are ``asked to make a bipolar judgment about whether a statement accurately describes a particular situation alluded to in some context or preamble''
% %; and are assumed to have 
% \citep{mcdaniel_truth-value_1996}. 
% %
% TVJ tasks assume the subject has 
% ``some conception of the notion of truth in the sense of a correspondence between what is said and the situation referred to'' \citep{mcdaniel_truth-value_1996}.
% Using this assumption, the subjects are then asked questions to probe their understanding of various grammatical constructions.

% We use TVJ tasks to explore if LLMs have similar notions of truth, specifically, if LLMs represent the truth of sentences in a way that is sensitive to context. 
% The task could be posed the same way it would be posed to a child, asking questions and making inferences about the LLM's competencies based on its answers.
% However, by using {belief probes}, we can infer its ``answer'' directly from the way it represents the input and learn how it changes throughout its layers. 
%
To probe for truth-value judgment we have a setup as displayed in \autoref{fig:setting_overview}, for example:\\[1mm]
\begin{tabular*}{\linewidth}{>{\hspace{12mm}}l>{\hspace{2mm} \small}l}
$Q$ \quad (premise)    & \text{\small \Xp{}.}
\\
$H$ \quad (hypothesis) & \text{\small \Xh{}.}
\end{tabular*}\\[1mm]
% The model input consists of a premise $Q$ and a hypothesis $H$.
The bracketed parts in $Q$ and $H$ are omitted or included to produce affirmed ($Q^+, H^+$) and negated ($Q^-, H^-$) sentences. 
We use truth-value probes to see if the model represents $H^+$/$H^-$ as true or false, and how this changes when preceded by $Q^+$ or $Q^-$. 
%
% The setup is similar to a natural language inference task. 
% However, we do not directly evaluate a model on its ability to classify sentence pairs by their \textit{meaning relation}: $R \in \{e, c, n\}$ (entailment, contradiction or neutral).
% Instead, we measure if the model's truth-value judgments (as measured by the probes) are consistent with it being able to differentiate between the meaning relations.

% It's important to remember that in order for this to work we must ensure our results are not distorted by data contamination.

% Our methodology is designed to investigate the context sensitivity of belief probes. 
There are different ways in which a belief could interact with the context of a statement. 
% The three kinds of beliefs that we consider are:
We define three kinds of beliefs in the following way:
\begin{itemize}[leftmargin=1em, itemsep=0pt, topsep=0pt]
    \item \textit{prior beliefs}, independent of the context, given by $\PLM(H)$;
    \item \textit{conditional beliefs}, specially where the context is assumed to be truthful, given by $\PLM(H | q)$;
    \item \textit{marginal beliefs}, where the truth of the premise and hypothesis are modeled jointly, with the effect of the premise summed out, given by $\sum_\tau \PLM(H, Q{=}\tau)$.\footnote{We leave this expression unsimplified to emphasize the dependence on the joint distribution, which is what distinguishes marginal from prior beliefs.}
\end{itemize}
Each of these types of beliefs are candidates for what is measured by $p(\vec{h}; {q})$, a probe applied to LLM activations for a hypothesis $H$ when preceded by a premise $Q$. 
Note that there could also be beliefs in between conditional and marginal, where the context's truthfulness is not totally assumed, but still biased (e.g. towards being true rather than false).
%swhich we can think of as also assigning a probability to the context's truthfulness.
%
% Conditional beliefs, marginal beliefs, and beliefs in between the two, can all be said to be truth-value judgment, because they are valid ways of incorporating the context.

% \bptodo{
% \item A truth-value probe associates a declarative sentence X with a probability $P(X)$. 
% \item By virtue of being a probability, any norms that govern probabilities necessarily also apply to the outputs of truth-value probes. Thus, in this case the probability that a sentence is false and the probability that it is true, must sum to one.
% \item Moreover, in order to faithfully be able to state that such a probe has recovered not just any probability, but a judgment (or occurrent belief), some additional norms apply. 
% These are the conditions given by \citet{herrmann_standards_2025}: accuracy, uniformity, coherence, and use.
% }

\subsection{Evaluation}\label{ssec:evaluation}
%
% averaging the probabilities for the affirmed and negated variants. 
%
To evaluate if probes coherently incorporate in-context information, we include four error scores, each indicating to what degree probe outputs violate desired behavior. 
% \autoref{tab:error_scores} shows the error scores and the (in)equalities on which they are based.

We first define the \textit{premise effect} ($\mathit{PE}$) as the difference in probability assigned to the hypothesis when preceded with an affirmed premise and probability assigned to the hypothesis on its own: $\mathit{PE} = p(\vec{h}; q^+) - p(\vec{h})$. 
We call a method's mean absolute premise effect its \textit{premise sensitivity}.
A value close to zero for this metric would be consistent with a {prior belief}. 
% than  and contextual ({conditional} or {marginal) beliefs on the other.%
%\todo{Mention in experiments section that this requires looking at the difference between probes trained with and without premises present.}

% The error scores are measure the extent to which methods are self-consistent.
The effect of adding the in-context premise can differ in magnitude depending on which probing method we use.
In order to make the error scores of different methods comparable, we express the errors in multiples of the premise effect $\mathit{PE}$. 
This makes the error scores independent of the overall premise sensitivity of the probing method. 

The first two error scores, E1 and E2 (see \autoref{tab:error_scores}) are based on the fact that we expect the probabilities to depend only on factors that are actually capable of influencing the truth value of the hypothesis. 
Thus, these error scores are proportional to the absolute change in probability that occurs after having the hypothesis preceded by either: 1) a corrupted premise $\tilde{Q}$, or 2) an unrelated premise $Q'$. 
The truth value of both corrupted and unrelated premises are independent of the truth value of the hypothesis, which is why we want the equalities for E1 and E2 in \autoref{tab:error_scores} to hold.

E3 and E4 measure when probes fail to behave like \textit{conditional} and \textit{marginal} beliefs, respectively. 
Consider the example: Q=\textit{``December is during the winter for New York''} and H=\textit{``In New York, days are shortest in December''}. 

% E3
If the model assumes the premise is true (a conditional belief, $p(\vec{h}; q) \approx \PLM(H|q)$) when determining the probability for H, then either: 
(1) having the context say ‘Q is incorrect’ should decrease the probability of H, or 
(2) having the context say ‘Q is correct’ should increase the probability. 
This expectation is captured by E3. 
% The error is based on inclusive inequalities for two reasons:
% (1) because the information in the premise might have already been incorporated, and (2) because if a premise is negated its meaning relation could become neutral (instead of flipping between entailment and contradiction).
For the error score, we have:
$% \begin{align}
    \left( p(\vec{h}; q^-) - p(\vec{h}) \right) \cdot \mathit{PE}^{-1} 
=   (p(\vec{h}; q^-) - p(\vec{h})) / (p(\vec{h}; q^+) - p(\vec{h})) 
$. % \end{align}
We want the effect of a negated premise to be opposite of a positive premise: when the numerator is positive, we want the denominator to be negative and vice versa. 
Taking $max\{\cdot, 0\}$ of this fraction, we can isolate the cases where the numerator and the denominator have the same sign, which are the errors we want to capture in the score. 

% For E3, we assume the model treats the context as truthful, and thus should consider the premise false when negated (and true when affirmed).
% If the premise is negated, the original meaning relation either switches (between entailment and contradiction), or becomes neutral. 
% When the relationship switches the premise effect should be opposite as well (from increasing the probability, to decreasing, and vice versa). 
% But, if negation creates a neutral relationship, then the probability should be the same as when there is no premise.
% Together, this gives us the inclusive inequalities in the left column of \autoref{tab:error_scores}.

% E4
If the model instead bases itself on its own evaluation of the truth of Q (a marginal belief, i.e. it models Q and H jointly: $p(\vec{h}; q) \approx \sum_\tau \PLM(h, Q{=}\tau)$), then having ‘Q is incorrect’ or ‘Q is correct’ should not influence the probability of H at all. This expectation is captured by E4.
% For E4, if the language model bases itself on its own evaluation of the premise, then it should ignore whether the premise is affirmed or negated. 
In that case, the probability assigned to the hypothesis should be the same regardless of whether the premise is asserted or denied.

Because low scores for E3 and E4 indicate two equally valid types of truth-value judgment, we do not expect the score to be low for both. See \autoref{apx:error_scores} for additional details.

\begin{table}[hbt]
    \centering
    \let\mc\multicolumn    
    \let\mr\multirow
    \small
    \renewcommand{\arraystretch}{1.3}
    \vspace{1mm}
    \caption{Expected behavior and corresponding error scores. The subscript $e$ and $c$ indicate hypotheses entailed or contradicted by their premise. }
    \vspace{-3mm}
    \begin{tabular}{rAA}
    \toprule
        & \mc{2}{c}{(in)equality} & \mc{2}{c}{error score}
 \\ \midrule
    E1  & \PLM(h|\hspace{0.2em}\tilde{Q}\hspace{0.2em})  &=  \PLM(h)     
        & \hspace{1.25cm} | p(\vec{h};\tilde{q}\hspace{2.3mm}) &- p(\vec{h}) | \cdot |\mathit{PE}^{-1}| 
 \\
    E2  & \PLM(h|\hspace{0.1em}Q')         &= \PLM(h)
        & | p(\vec{h};q'\hspace{1.2mm}) &- p(\vec{h}) | \cdot |\mathit{PE}^{-1}| 
 \\
    \mr{2}{*}{E3} 
        % & \PLM(h_e |\neg q)        &\leq    \PLM(h)
        & \PLM(h_e |q^-)        \leq    P&_\lambda(h) \leq \PLM(h_e | q^+)
        & \mc{2}{r}{ \mr{2}{*}{ {
              \begin{minipage}{5.5cm}\vspace{-3mm}\begin{equation*}%
                \max\{ 
                    ( p(\vec{h};q^-) - p(\vec{h}) ) \cdot \mathit{PE}^{-1} ,
                \; 0 \} 
              \end{equation*}\end{minipage}%
            }}}
 \\
        & \PLM(h_c |q^-)        \geq    P&_\lambda(h) \geq \PLM(h_c | q^+)& &
 \\
    % E4  & \mathbb{E}_{\tau \sim p(\vec{q}^-)}[\PLM(h|Q{=}\tau)] 
           % & \approx \mathbb{E}_{\tau \sim p(\vec{q}^+)}[\PLM(h|Q{=}\tau)]
    % E4  & \mathbb{E}_{\PLM(Q^-)}[\PLM(h|Q^-)] 
           % & \approx \mathbb{E}_{\PLM(Q^+)}[\PLM(h|Q^+)]
    E4  & \sum_\tau \PLM(h, Q^-{=}\tau)
        & = \sum_\tau \PLM(h, Q^+{=}\tau)
        & | p(\vec{h};q^-) &- p(\vec{h};q^+) | \cdot |\mathit{PE}^{-1}|
 \\ \bottomrule
    \end{tabular}
    \label{tab:error_scores}
\end{table}

\section{Experiments}%
% \todo{add note: the different role that the polarity plays for the hypothesis and the premise}%
In our experiments, we make use of datasets with samples of related sentences whose truth values depend on each other.
We use samples from these datasets by creating prompts where the sentences are either affirmed or negated.

% \subsection*{Training \& evaluating probes}
We train probes in a \texttt{no-prem} and \texttt{pos-prem} setting.
For \texttt{no-prem}, the premise $Q$ is left out, and for \texttt{pos-prem} the premise appears in the positive (or affirmed) variant.
We include these settings to better understand how truth-values are represented.
A direction found in the \texttt{no-prem} setting we might expect to represent {prior belief}. 
If that direction shows context-sensitivity (when evaluated with premises in-context), that is evidence that the model does not represent the prior and contextual beliefs independently (in orthogonal directions). 
For \texttt{pos-prem}, the direction found is also influenced by what appears in context. 
If the directions found for \texttt{pos-prem} and \texttt{no-prem} are different, it suggests there \textit{is} a separate (but possibly related) direction used to represent contextual belief.

% The probe inputs $\vec{h}$ are the representations of the answer tokens (`correct' / `incorrect'), extracted for each layer.
The probe inputs $\vec{h}$ are the representations of the period following the answer tokens (`correct' / `incorrect') extracted for each layer.
To compare across probing methods we calibrate the probes such that their predictions for the $p(\vec{h})$ case have the same variance. 
We train probes on the following LLMs: Llama2-7b, Llama2-13b \citep{touvron_llama_2023b}, and 
OLMo-7b 
with and without instruction tuning \citep{groeneveld_olmo_2024}. 

%%%%%%%%%%%%%%%%%%%%
%%%% EVALUATION %%%%
%%%%%%%%%%%%%%%%%%%%
To measure the premise effect, and error scores described in \autoref{ssec:evaluation}, we include the following evaluation cases:  
$p(\vec{h})$ (no premise),
$p(\vec{h}; {q}^+)$ (affirmed premise),
$p(\vec{h}; {q}^-)$ (negated premise),
$p(\vec{h}; {q'})$ (unrelated premise), and
$p(\vec{h}; \tilde{q})$ (corrupted premise).
We evaluate both the \texttt{no-prem} and \texttt{pos-prem} in all of these cases. The first two cases are `in distribution' for the \texttt{no-prem} and \texttt{pos-prem} settings, respectively. 
The other combinations are out of distribution.
When evaluating the probes we use: $p(\vec{h}) = \frac{1}{2}(1 - p(\vec{h}^-) + p(\vec{h}^+)).$

\subsection*{Data}\label{ssec:experiments:data}
We use two existing datasets in our experiments. 
The first dataset \citep[EntailmentBank, ][]{dalvi_explaining_2021} contains hypotheses that are sentences with general world knowledge. These are facts the LLM may have encountered during training and for which it could already have a strong prior belief.
The second \citep[SNLI, ][]{bowman_large_2015} contains statements that describe images, to which an LLM has no access.
For both datasets, the corrupted sentences are created by replacing the characters in each word of the base sentence with random characters.
% In the examples below, the bracketed characters determine if the premise and hypothesis are positive or negative. 
%
The polarity of the premises and hypotheses are determined by switching between sentences that say something is `correct' and saying that it is `incorrect'. 
This style of negation avoids some problems that might otherwise arise.\footnote{For example, negating ``four children are playing in some water'' as ``four children are not playing in some water'', still presupposes the existence of four children. Using a negative meta statement leaves open the possibility that the presupposition is false (e.g. the number of children is inaccurate).}

\paragraph{EntailmentBank} This dataset contains statements with entailment relationships. 
% This dataset is similar in structure to SNLI, consisting of premises and hypotheses, but it contains only entailments.
The dataset was derived from ARC \citep{clark_think_2018}, which consists of grade-school level science questions.
We combine premises from EntailmentBank with the questions and answers from ARC.
The questions are answered correctly or incorrectly to create both entailments and contradictions. For example:\\[1mm]
\begin{tabular*}{\linewidth}{>{\hspace{5mm}}l>{\small}l}
        & You are given the following question:
\\
        & $>$ In New York, the shortest period of daylight occurs during? (A) December (B) June
\\
$Q_{a}$ & The statement ``New York is located in the northern hemisphere.'' is [in]correct.
\\
$Q_{b}$ & The statement ``December is during the winter for New York.'' is [in]correct.
\\
$H$     & Answering the question with ``(B) June'' is [in]correct.
\end{tabular*}\\[1mm]
%\begin{align*}
           % \quad & \text{\small You are given the following question:}\\
           % \quad & \text{\small $>$ In New York, the shortest period of daylight occurs during? (A) December (B) June}\\
    % Q_{a}  \;\;  & \text{\small The statement ``New York is located in the northern hemisphere.'' is [in]correct.}\\
    % Q_{b}  \;\;  & \text{\small The statement ``December is during the winter for New York.'' is [in]correct.}\\
    % H      \quad & \text{\small Answering the question with ``(B) June'' is [in]correct.}
% \end{align*}%
The answer ``June'' is incorrect, and thus $H$ contradicts the information in $Q_a, Q_b$ (when it is not negated), while in the sample with the correct answer $H$ would be entailed by $Q_a, Q_b$.
The dataset contains trees of entailing sentences, but we disregard anything but the first level of supporting premises.
For the $p(\vec{h}; q)$ case, we use the distractor premises provided in the dataset. 
These were ranked as potentially relevant, but during annotation were not selected to be part of the entailment tree \citep{dalvi_explaining_2021}. 
%Another difference is that, each premise on its own does not necessarily have an entailment relation with the hypothesis, but their conjunction does. 

\paragraph{SNLI}
This dataset is a Natural Language Inference dataset, it consists of premise-hypothesis pairs, which are labeled as: entailment, contradiction, or neutral; describing the meaning relation between the sentences.
This dataset was created based on the descriptions of images. 
To avoid ambiguity, we establish a context as follows: \\[1mm]
\begin{tabular*}{\linewidth}{>{\hspace{5mm}}l>{\small}l}
        & You are looking at a picture (A) which is placed next to an unrelated picture (B).
\\
$Q$     & Describing picture \{A/B\} as: ``Four children are playing in some water.'' is [in]correct.
\\
$H$     & Saying (about picture A) that: ``The children are wet.'' is [in]correct.
\end{tabular*}\\[1mm]
% \begin{align*}
%         \quad & \text{\small You are looking at a picture (A) which is placed next to an unrelated picture (B).}\\
%     Q   \quad & \text{\small Describing picture \{A/B\} as: ``Four children are playing in some water.'' is [in]correct.}\\
%     H   \quad & \text{\small Saying (about picture A) that: ``The children are wet.'' is [in]correct.}
% \end{align*}
The neutral sentences for the $p(\vec{h}; q')$ case are obtained by taking the premise from a different, randomly sampled premise-hypothesis pair.
Furthermore, for this case, the `A/B' that appears in curly brackets is set to B to ensure that there is a fully neutral relationship. 
Without it, the fact that the two sentences are about the same picture could make their (simultaneous) truth less likely.
It is also set to B for $p(\vec{h}; \tilde{q})$, and set to A for all other cases.

Without access to the picture, the model's prior belief should result in 50\% accuracy.
However, for SNLI it is possible to predict the label solely from the hypothesis \citep{poliak_hypothesis_2018}.
This makes for an interesting scenario when it comes to truth-value probing.
A probing method might identify a direction that only encapsulates a statistical pattern, rather than a model's truth-value direction.
It is also possible that the statistical pattern is absorbed into the model's truth-value direction, as simply another reason to believe the sentence.
After the addition of a premise, we do not expect a representation should move (coherently) in a direction which merely encodes a statistical pattern.
Thus, if a probe trained only on hypotheses \textit{does} respond coherently to the presence of a premise at test time, it is further evidence of the probe uncovering truth-value judgment, and not just a statistical pattern.

        % % % % % % % % % % % % % 
 % % % %                         % % % %
%              EXPERIMENT 1             %
 % % % %                         % % % %
        % % % % % % % % % % % % %
\subsection{The effects of altering premises}
We evaluate the probes on held-out data, including data from all the other variants. 
We also include an additional baseline, based on the model's LM-head, where the probabilities assigned to the `correct'/`incorrect' tokens are rescaled to sum up to one.%

% \subsubsection*{Results}
\autoref{tab:results} gives an overview of the average probabilities for $p(\vec{h}; q^+)$, $p(\vec{h}; q^-)$, and $p(\vec{h})$, split by whether the premise-hypothesis pair had an entailment or contradiction relation. 
% For each method there are two rows, one for the layer where the probe obtained the best accuracy, and another where the probe obtained the lowest error scores. 
We observe that the \emph{probabilities assigned to hypotheses depend strongly on the presence of relevant premises}.
When the hypothesis is entailed the probabilities are higher, and when the hypothesis is contradicted they are lower.
This is true, even for probes trained without the premises present (\texttt{no-prem}), which also achieve good accuracy for the $p(\vec{h}; q^+)$ case. 
Although the premise sensitivity is lower, it is clear that the directions identified in the \texttt{no-prem} setting are not encoding specifically \textit{prior} beliefs.
% Our results indicate that the belief direction derived from non-contextual data is still shows premise sensitivity when evaluated on data with premises.
% Note that the accuracy of the probes trained without premises also increases, higher then when evaluated on data without premises, indicating the 

\def\g{\color{gray}}

\begin{table}[ht!]%
\centering%
% \footnotesize%
% \scriptsize%
\semiscript%
% \fontsize{7}{7}\selectfont
\renewcommand{\arraystretch}{0.8}%
\setlength\extrarowheight{4pt}%
\setlength\tabcolsep{3pt}%
% \vspace{-2mm}
\caption{Accuracy of $p(\vec{h}; q^+)$ (Acc), mean probabilities (orange=0, gray=0.5, blue=1), and trimmed mean errors scores for probes of each method on both datasets for Llama2-7b. The probes are from layers (L) with: (1) the best accuracy; and (2) the overall lowest error scores (by average error rank $E*$). The best scores per dataset are in bold, for E3 and E4 the bold values are based on their sum. CCS omitted, full table in \autoref{apx:additional_tables}.}\vspace{-2mm}%
\begin{tabular}{lllccrRRRRRrrrr}%
    \addlinespace[-1mm]\toprule\addlinespace[0.2mm]%
       &&&&&& \multicolumn{2}{c}{Entailment} & \multicolumn{1}{c}{} & \multicolumn{2}{c}{Contradiction} \\
    \addlinespace[0mm]\cmidrule{7-8}\cmidrule{10-11}\addlinespace[-1mm] 
       & & Method & L & Acc & \multicolumn{1}{c}{$E*$} &
        \multicolumn{1}{c}{$p(\vec{h}; q^+\hspace{-0.3mm})\hspace{-0.3mm}$} &  
        \multicolumn{1}{c}{$p(\vec{h}; q^-\hspace{-0.3mm})\hspace{-0.3mm}$} &  
        \multicolumn{1}{c}{$p(\vec{h})$} &  
        \multicolumn{1}{c}{$p(\vec{h}; q^-\hspace{-0.3mm})\hspace{-0.3mm}$} &  
        \multicolumn{1}{c}{$p(\vec{h}; q^+\hspace{-0.3mm})\hspace{-0.3mm}$} 
        & E1 & E2 & E3 & E4 \\ 
    \midrule[\heavyrulewidth]\addlinespace[0mm]
    \multirow{13}{*}{\rotatebox[origin=c]{91}{EntailmentBank}}
&&LM-head& -  &     .80 &    145.8 & .61 & .52 & .50 & .49 & .38 &     .96 &     .90 &     .31 &     1.11 \\
   \addlinespace[0mm]\cmidrule{2-15}\addlinespace[-1mm]   
&  \multirow{6}{*}{\rotatebox[origin=c]{90}{\texttt{no-prem}}}   
 & CCR   & 14 &    .63 & \g  141.4 & .55 & .52 & .49 & .48 & .45 &    1.04 &    1.22 &     .99 &      .62 \\
&&       & 29 & \g .58 &     127.4 & .53 & .51 & .49 & .48 & .46 &     .93 &    1.17 &     .86 &      .74 \\
&& LR    & 16 &    .93 & \g  160.0 & .78 & .59 & .50 & .41 & .24 &    1.04 &     .90 &     .21 &     1.36 \\
&&       & 14 & \g .92 &     107.6 & .75 & .61 & .50 & .39 & .25 &     .89 &     .85 &     .28 &     1.15 \\
&& MMP   & 19 &    .89 & \g  145.2 & .71 & .54 & .49 & .46 & .31 &     .68 &     .79 &     .20 &     1.28 \\
&&       & 22 & \g .86 &     103.6 & .69 & .53 & .49 & .47 & .33 &     .71 &     .83 &     .31 &     1.17 \\ 
   \addlinespace[0mm]\cmidrule{2-15}\addlinespace[-1mm]   
&  \multirow{6}{*}{\rotatebox[origin=c]{90}{\texttt{pos-prem}}}   
 & CCR   & 16 &     .87 & \g  89.0 & .86 & .54 & .50 & .46 & .18 &     .56 &     .67 &     .05 &     1.27 \\
&&       & 14 & \g  .86 &     70.0 & .84 & .52 & .50 & .49 & .18 &     .57 &     .65 &     .05 &     1.27 \\
&& LR    & 18 & \bf .96 & \g  51.6 & .92 & .60 & .50 & .40 & .10 &     .52 &     .58 & \bf .08 & \bf 1.16 \\
&&       & 14 & \g  .95 & \bf 43.6 & .91 & .60 & .49 & .41 & .11 & \bf .43 & \bf .56 & \bf .08 & \bf 1.16 \\
&& MMP   & 14 &     .89 & \g  60.6 & .86 & .52 & .50 & .49 & .16 &     .51 &     .61 &     .04 &     1.26 \\
&&       & 14 & \g  .89 &     60.6 & .86 & .52 & .50 & .49 & .16 &     .51 &     .61 &     .04 &     1.26 \\ 
   \addlinespace[0mm]\midrule[\heavyrulewidth]\addlinespace[0mm]   
   \multirow{13}{*}{\rotatebox[origin=c]{90}{SNLI}}   
&&LM-head & -   & .62  &  150.6   & .57 & .54  & .52  & .43  & .43  & .89  &  .88  &  .36 &  1.35 \\
   \addlinespace[0mm]\cmidrule{2-15}\addlinespace[-1mm]   
& \multirow{6}{*}{\rotatebox[origin=c]{90}{\texttt{no-prem}}}   
& CCR     & 7   &    .57  & \g 138.8  & .52  & .52  & .53  & .49  & .49  & .93  & 1.02  & 1.16 &  .26 \\
&&        & 12  & \g .52  &    100.2  & .51  & .53  & .51  & .47  & .50  & .74  &  .95  &  .99 &  .27 \\
&&LR      & 13  &    .85  & \g 189.8  & .67  & .75  & .50  & .24  & .32  & .91  & 1.13  &  .89 & 1.13 \\
&&        & 20  & \g .75  &    103.4  & .65  & .57  & .50  & .42  & .35  & .72  &  .96  &  .37 & 1.21 \\
&&MMP     & 13  &    .88  & \g 178.2  & .61  & .65  & .50  & .35  & .38  & .91  & 1.06  & 1.03 &  .54 \\
&&        & 32  & \g .45  &    129.0  & .48  & .51  & .51  & .49  & .52  & .92  & 1.04  &  .68 &  .87 \\
\addlinespace[0mm]\cmidrule{2-15}\addlinespace[-1mm]
&  \multirow{6}{*}{\rotatebox[origin=c]{90}{\texttt{pos-prem}}}
& CCR     & 26  & .91     & \g  53.8   & .87  & .68  & .50  & .28  & .14  &     .42  &     .53  &  .47 &   .60 \\
&&        & 28  & \g  .91 &     53.6   & .86  & .70  & .50  & .28  & .14  &     .41  &     .51  &  .49 &   .57 \\
&&LR      & 16  & \bf .95 & \g  95.6   & .93  & .77  & .51  & .22  & .06  &     .47  &     .61  &  .63 &   .42 \\
&&        & 26  & \g  .95 & \bf 41.8   & .88  & .68  & .50  & .29  & .11  & \bf .38  & \bf .48  &  .44 &   .61 \\
&&MMP     & 17  &     .94 & \g  90.0   & .92  & .77  & .50  & .20  & .09  &     .46  &     .57  & \bf .68 & \bf .35 \\
&&        & 6   & \g  .74 &     49.6   & .69  & .65  & .50  & .34  & .27  &     .39  &     .50  &  .62 &   .44 \\
    \addlinespace[0mm]\bottomrule[0.8pt]\addlinespace[-1mm]
\end{tabular}%
\label{tab:results}%
\end{table}%

\begin{figure}[ht!]%
    \vspace{-3mm}
    \begin{subfigure}{0.50\textwidth}%
        \includegraphics[
            height=1.6in, trim=0.1in 0 0 2mm, clip
            % trim=left bottom right top
        ]{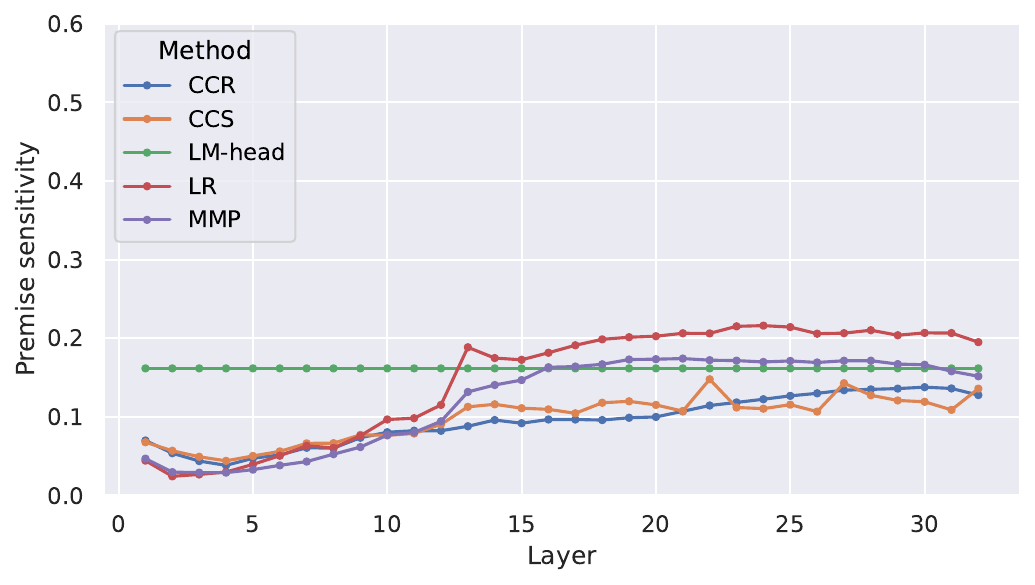}%
    \vspace{-3mm}
    \caption{Trained on \texttt{no-prem}}%
    \end{subfigure}%
    \begin{subfigure}{0.50\textwidth}
        \includegraphics[
            height=1.6in, trim=0.26in 0 0 2mm, clip
        ]{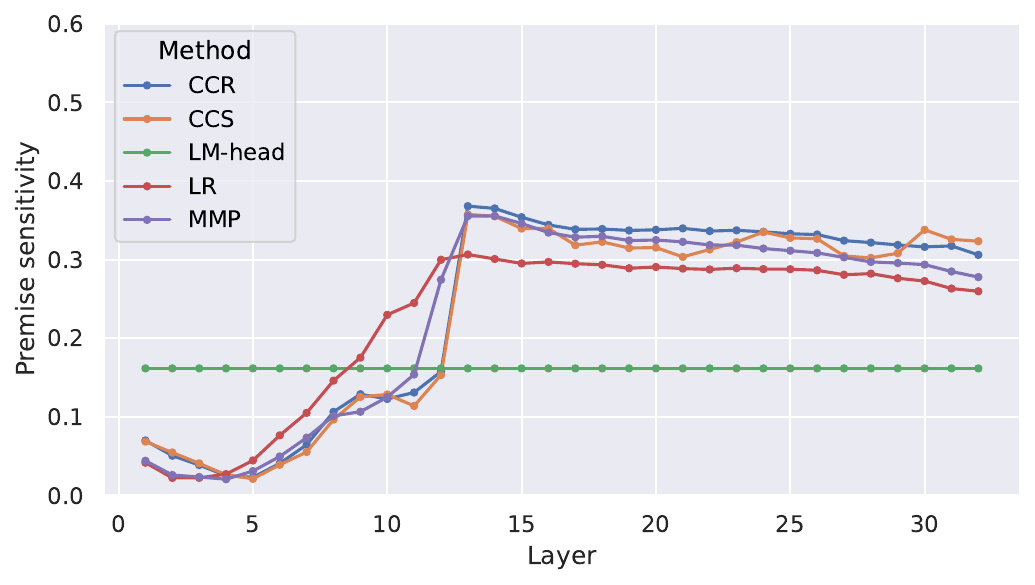}%
        \vspace{-3mm}
        \caption{Trained on \texttt{pos-prem}}%
    \end{subfigure}\vspace{-3mm}%
    \caption{Premise sensitivity for Llama2-7b on EntailmentBank.}%
    \label{fig:premise_sensitivity}%
\end{figure}%
\begin{figure}[ht!]%
    \vspace{-1mm}%
    \begin{subfigure}{0.50\textwidth}%
        \includegraphics[
            height=1.43in, trim=0.1in 0 0 12mm, clip
            % trim=left bottom right top
        ]{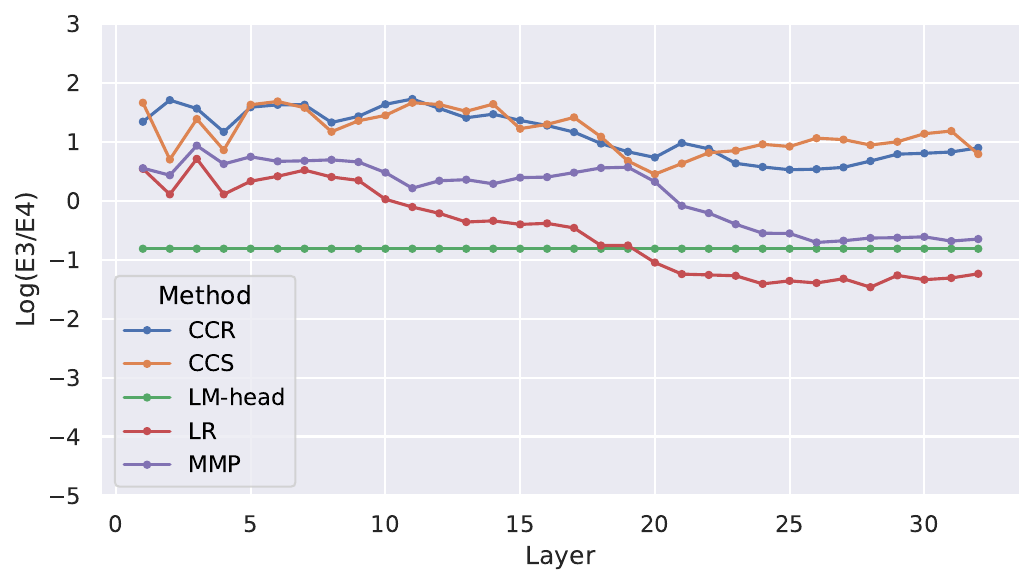}%
        \vspace{-2.5mm}
        \caption{OLMo-7B}%
    \end{subfigure}%
    \begin{subfigure}{0.50\textwidth}
        \includegraphics[
            height=1.43in, trim=0.26in 0 0 12mm, clip
        ]{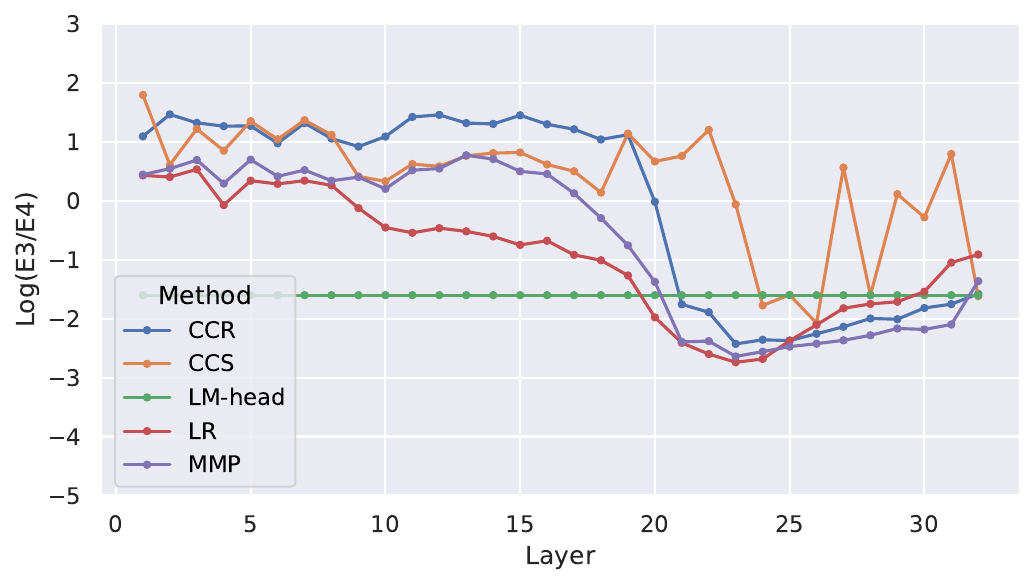}%
        \vspace{-2.5mm}
        \caption{OLMo-7B-Instruct}\label{fig:e3-e4:olmo_instr}%
    \end{subfigure}\vspace{-3mm}%
    \caption{Log-ratio of E3 and E4 error score for probes trained using the \texttt{no-prem} variants of EntailmentBank on OLMo-7B and OLMo-7B-Instruct.}
    \label{fig:e3-e4-olmo-vs-olmo_instr}
\end{figure}

\vspace{-5mm}\paragraph{Error scores.} 
In \autoref{tab:results}, we can see that especially E1 and E2 are quite high.
This suggests that \emph{the identified directions are sensitive to irrelevant information}.
Probes trained on \texttt{no-prem} often have E1 and E2 close to one.
Because the error scores are normalized by the premise effect, a value of one means that, on average, a corrupted or unrelated premise has an effect with the same magnitude as the original affirmed premise.
The error scores improve when probes are trained on \texttt{pos-prem}.
Comparing Llama2-7b to Llama2-13b (see \autoref{tab:llama2-13b_results}) shows the scores are not consistently lower for the larger model. 
% meaning \emph{error scores show no sign of scaling with model size.}
%

\paragraph{LM-head baseline.}
Most probes beat the LM-head in terms of accuracy and premise sensitivity. 
This suggests that \emph{inconsistency can occur even when the LLM’s representations contain information able to prevent it}. 
This is in line with findings for LLM hallucinations.

\paragraph{Premise sensitivity by layer.}
\autoref{fig:premise_sensitivity} shows the premise sensitivity across layers for probes of each method when applied to Llama2-7b. These were trained on the \texttt{no-prem} (left) and \texttt{pos-prem} (right) variants of the EntailmentBank data.
All methods show a degree of premise sensitivity, with \texttt{no-prem} showing less than \texttt{pos-prem}.
There do not seem to be layers where the probe is not sensitive to the premises (approximating $\PLM(H)$), while still having above random accuracy (see \autoref{apx:prem_acc}). 
Suggesting that \emph{LLMs do not represent prior beliefs $\PLM(H)$ fully independently}.

\paragraph{Pretrained-only vs. instruction-tuned.}
\autoref{fig:e3-e4-olmo-vs-olmo_instr}
%\todo{note on disagreement}\todo{something about CCS's variance} shows the log ratio of E3 and E4. 
In the later layers of the instruction-tuned model, it leans more toward E4 errors. 
The instruct-tuned model's behaviour is a lot more sensitive to whether the premise is negated or affirmed.
This suggests that \emph{instruction-tuning makes the model more likely to represent prior assertions as true}, which is consistent with the instruction-tuning objective. 
%\todo{Expand on this a bit. Does this mean the "prior knowledge" is overridden more easily by in-context assertions? }

\paragraph{Spurious correlations.} 
Looking at SNLI, both LR and MMP show premise sensitivity, suggesting that they find directions indicative of more than just the spurious correlations present in the hypotheses of SNLI.
However, for LR the probe's behaviour does seem affected by the spurious correlations.
Its average probabilities for samples with negated premises is not between the probabilities obtained for samples with positive premises and no premises, resulting in a high E3+E4 score.

% \begin{figure}%
%     \centering%
%     \begin{subfigure}[t]{0.49\textwidth}%
%         \centering%
%         \includegraphics[
%             height=1.5in, trim=0 0 0 0, clip
%         ]{figures/snli/llama2-7b/log_ratio_E3_E4plot_train_no-prem_Llama-2-7b-hf.pdf}%
%     \caption{SNLI}%
%     \end{subfigure}%
%     \begin{subfigure}[t]{0.49\textwidth}%
%         \centering%
%         \includegraphics[
%             height=1.5in,
%         ]{figures/entbank/llama2-7b/log_ratio_E3_E4plot_train_no-prem_Llama-2-7b-hf.pdf}%
%         \caption{EntailmentBank}%
%     \end{subfigure}%
%     \caption{The log-ratio of E3 to E4, showing which error dominates in each layer.}%
%     \label{fig:log_ratio_e3e4}%
% \end{figure}
% In \autoref{fig:log_ratio_e3e4} we can see the log-ratio of E3 to E4 for Llama2-7b on both datasets. It shows which of the two complementary errors dominates in each layer. 
% What we see for SNLI is that in the early layers E4 is larger than E3 for all probing methods. 
% This suggests that in the early layers what is stated about the premises is assumed to be true. 
% %In later layers E3 and E4 are closer in value, suggesting the model  
% On EntailmentBank this does not happen, which can be explained through the differences between the two datasets: the EntailmentBank hypotheses state facts for which LLMs likely already have degree-of-belief regardless of the premise.

        % % % % % % % % % % % % % 
 % % % %                         % % % %
%              EXPERIMENT 2             %
 % % % %                         % % % %
        % % % % % % % % % % % % %
\subsection{Intervening on premise representations}%
\begin{wrapfigure}[13]{r}{0.5\textwidth}%
    \vspace{-5.5mm}
    \hspace{-1mm}
    \centering
    \includegraphics[
        width=0.5\textwidth,
        %trim=left bottom right top
        trim=2mm 0mm 2mm 3mm, clip
    ]{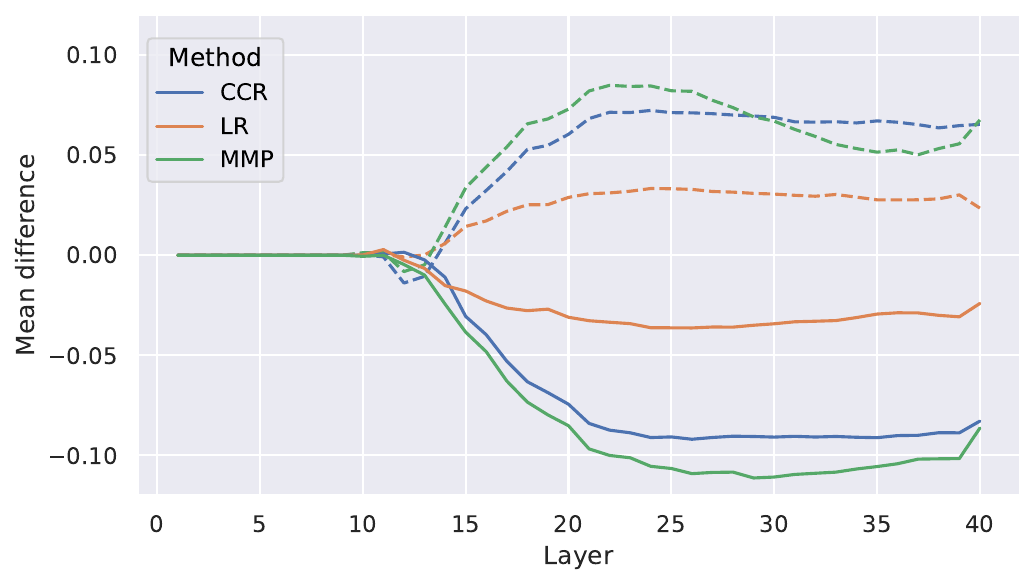}%
    \llap{\raisebox{0.75cm}{\includegraphics[
        width=1.5cm, clip
    ]{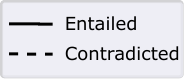}}\hspace{4.55cm}}%
    \vspace{-3mm}%
    \caption{Intervention effect: mean difference in probability $p(\vec{h}; do(\vec{q}^+{\mathrel{-}=}\bm{\theta})) - p(\vec{h}; q^+)$ by layer, for entailments and contradictions.}
    \label{fig:causal_result}
\end{wrapfigure}
In this experiment, we alter the LLM's internal representations directly, rather than only altering the input data. 
We take the directions found by the probing methods in the first experiment, and move the representations of the premises along this direction.

We perform this experiment for the $p(\vec{h}; q^+)$ and $p(\vec{h}; q^-)$ cases on the EntailmentBank data. 
% We evaluate the probability of the hypothesis using the same direction as the premise was moved in. 
We move the premise in the direction found during \texttt{pos-prem} training, and use that same direction to evaluate pre-intervention: $p(\vec{h}; q)$, and post-intervention: $p(\vec{h}; do(\vec{q}{ \mathrel{\pm}=}\bm{\theta}))$. 
We perform the intervention using the same method and parameters as \citet{marks_geometry_2024}.
The intervention is done on Llama2-13b in layers 8-14, and applied to the representations of the answer tokens (correct, incorrect), and the period after.
All interventions have the same magnitude: $|\bm{\theta}_\text{mm}|$.

\paragraph{Results.}
In \autoref{fig:causal_result}, we can see the effect of the causal intervention for the $p(\vec{h}; q^+)$ case.
When we move the affirmed premises backwards in the truth-value direction, the probabilities of entailed and contradicted hypotheses decrease and increase, respectively.
This shows that \emph{truth-value directions causally mediate the incorporation of in-context information}.
We see that intervening with the direction found by LR has a smaller effect than MMP and CCR.
The largest change is a reduction of around ten percentage points for entailed hypotheses. 
See \autoref{fig:causal_neg} for the results of $p(\vec{h}; do(\vec{q}^-{\mathrel{+}=}\bm{\theta}))$.

\section{Discussion \& Conclusion}
We have investigated LLM truth-value judgment, which requires correctly incorporating context when determining the truth value of a sentence.
Based on our expectations of how the probability of a sentence should or should not change in a supporting, contradicting, or neutral context, we created four error scores.
In our experiments, we used several probing methods on four language models, and quantified how they assign probabilities to hypotheses in different contexts.
From our results, the following becomes clear:
\begin{itemize}[leftmargin=1.5em, itemsep=0pt, topsep=0pt]
\item 
LLMs incorporate context when representing sentences as more or less (likely to be) true. 
However, contexts which should have no bearing on truth values still have a sizable impact on a sentence's position along the direction identified by the probes.
Thus, in our evaluation the LLMs exhibited only a limited degree of \textit{coherence}, suggesting attributions of belief to LLMs based on current truth-value probes are unfaithful. 

\item The positioning of premises along truth-value directions partially determines the positioning of related hypotheses along the same direction. This shows the directions are causal mediators of the inference process, which is likely part of a mechanism that, when fully uncovered, will help explain how LLMs tackle reasoning tasks. 
\item Among the tested truth-value probing methods, Logistic Regression failed to reveal levels of coherence and causal mediation that the others did, showing its limitations.
\end{itemize}

In principle, the lack of coherence can be attributed to both flaws in the probes or flaws in the model. 
However, we include multiple probing methods, thereby mitigating the risk that the results are due to a particular flaw in any single probing method. 
The sensitivity to irrelevant information we report is also consistent with previous black-box evaluations \citep{shi_large_2023}.
% Our methodology allows the community to evaluate truth-value probes in new ways. 
If the probes are at fault, then our methodology provides new ways of evaluating, helping to distinguish between good and bad truth-value probes.

% The overall higher error scores shows that the probes direction found by training probes on 
% The sensitivity to irrelevant information observed here is consistent with previous findings from \citep{shi_large_2023}. 

Our findings further show the existence of separate (albeit possibly related) directions that can be found depending on whether the probes are based on inputs consisting of individual sentences (\texttt{no-prem}), or sentences that occur in a context (\texttt{pos-prem}).
% In our experiments we have investigated one direction at a time.
Recently, \citet{burger_truth_2024} showed that truth-values in LLMs occupy a two-dimensional subspace: 
one direction consistently points from true to false, and 
another is polarity-sensitive and points from false to true for negated statements.
Future work should investigate whether there is a subspace which encodes truth-values in different ways, corresponding more closely to either prior, marginal or conditional beliefs. 
% However, finding them will require either new probing methods or the construction of very specific data where the `\textit{being entailed/contradicted by context}' and `\textit{being true/false}' features can be varied completely independently. We leave this for future work.

Future work should also seek to better understand the representations of meaning-relations in LLMs, and the exact mechanisms responsible for incorporating that information into the truth-value directions.
For example, by investigating the construction of probes that reveal if a model represents two sentences as having a particular meaning relation. 
% Then, we can detect when the model disagrees with the gold standard meaning relation provided by the dataset. 
% With probabilities for all three relevant variables: $H$, $Q$, $R$, an even more precise evaluation would become possible.

% In our experiments we have only investigated models with 7 or 13 billion parameters. 
% To fully investigate the interaction of our error scores with model size, additional experiments are needed.

\FloatBarrier

\section*{Acknowledgments}
This research was supported by Huawei Finland through the DreamsLab project. All content represented the opinions of the authors, which were not necessarily shared or endorsed by their respective employers and/or sponsors.

% \section*{Ethics Statement}
% Authors can add an optional ethics statement to the paper. 
% For papers that touch on ethical issues, this section will be evaluated as part of the review process. The ethics statement should come at the end of the paper. It does not count toward the page limit, but should not be more than 1 page. 

\bibliography{references2}
\bibliographystyle{colm2025_conference}

\appendix

\renewcommand\thefigure{\thesection.\arabic{figure}}    
\renewcommand\thetable{\thesection.\arabic{table}}    
\setcounter{figure}{0}
\setcounter{table}{0}
\newpage
\section{Error scores}\label{apx:error_scores}
\begin{wrapfigure}{r}{6cm}%
    \centering%
    \includegraphics[width=5.5cm]{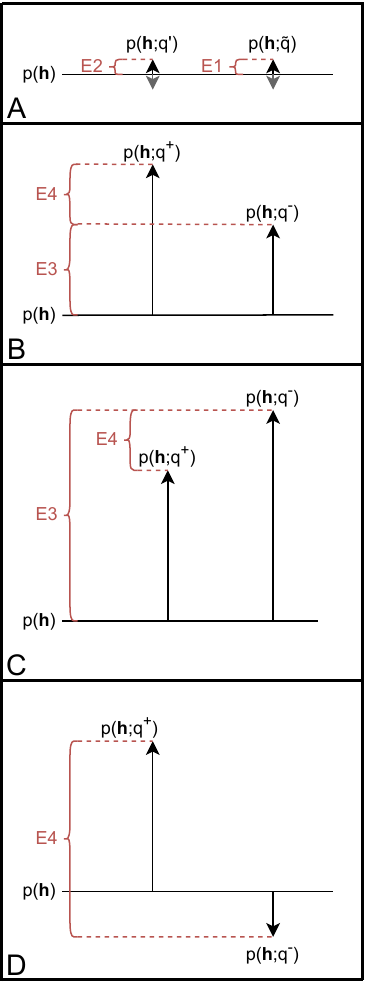}%
    \caption{Error score diagram.}%
    \vspace{-1cm}%
    \label{fig:consistency_aid}%
\end{wrapfigure}

%\todo{Revise, focus on geometric intuition}
% Our consistency error scores are explained in more detail here.
Here we try to give some (geometric) intuitions for our error scores.
% We go over different scenarios and show which errors are involved.
Specifically, we make use of the diagrams presented in \autoref{fig:consistency_aid}.
These diagrams take as a baseline the probability assigned to the hypothesis on its own $p(\vec{h})$, and show all other probabilities relative to it. 
The diagram assumes we are looking at premise-hypothesis pairs with entailment relations. 
The diagrams for contradictions would be identical, but mirrored vertically.

E1 and E2 consistency errors are shown in box A in \autoref{fig:consistency_aid}. 
Both of these errors involve the difference in probability assigned to (a) the hypothesis on its own and (b) the hypothesis preceded with an irrelevant statement, which is either:
\begin{itemize}[leftmargin=7mm, noitemsep, topsep=-1mm]
    \item a premise where the characters have been replaced by random characters $p(\vec{h}; \tilde{q}$); or
    \item a premise that has been replaced by another randomly sampled premise $p(\vec{h}; q')$. 
\end{itemize}
See \autoref{apx:data_samples} for examples.

E3 and E4 consistency errors are indicative of two opposing behaviours potentially exhibited by a language model. 
E3 assumes that the context (containing the premise) is truthful, and that what is asserted should be taken at face value.
If a contradicting premise is (said to be) true this should reduce the probability assigned to the hypothesis, and if a supporting premise is (said to be) true it should increase the probability assigned to the hypothesis.
On the other hand, E4 is assumes that the model uses its own evaluation of the context, ignoring if it is asserted to be true or false. 
If this is the case, then the probability assigned to the hypothesis should not depend on the truth value that is asserted of the premise.
These two are displayed in three different scenarios (B, C, D) in \autoref{fig:consistency_aid}.

In B, we have $p(\vec{h}) < p(\vec{h};q^-) < p(\vec{h};q^+)$, in this scenario it is always the case that $E3 + E4 = 1$ (recall that the error scores are given as multiples of $\mathit{PE} = p(\vec{h};q^+) - p(\vec{h})$).
When evaluating the overall consistency of the model this is the best score for $E3 + E4$ that we can expect.

In C, we have $p(\vec{h}) < p(\vec{h};q^+) < p(\vec{h};q^-)$, this scenario is `double wrong', in that there is now a part of the probability that is punished by both error scores. 
Regardless of whether the model trusts that the context is truthful or trusts itself, it should never give a higher probability to an entailed hypothesis after seeing the premise negated than when it saw it affirmed.

In D, we have $p(\vec{h};q^-) < p(\vec{h}) < p(\vec{h};q^+)$, now we have E3 equal to zero, since it is perfectly acceptable for the probability of the hypothesis to decrease when preceded by a negated supporting premise. This can occur in two ways, either the supporting premise became a contradicting premise and thus makes the hypothesis less likely, or the premise became neutral, in which case it still takes away one (potentially important) reason to believe the hypothesis.
\setcounter{figure}{0}
\setcounter{table}{0}

\newpage

\section{Additional Tables}\label{apx:additional_tables}

\subsection{Llama2-7b}
\def\g{\color{gray}}
\begin{table}[h]%
\centering%
\footnotesize%
\renewcommand{\arraystretch}{0.8}%
\setlength\extrarowheight{4pt}%
\setlength\tabcolsep{3pt}%
\begin{tabular}{lllccrRRRRRrrrr}%
    \addlinespace[-1mm]\toprule\addlinespace[0.6mm]%
       &&&&&& \multicolumn{2}{c}{Entailment} & \multicolumn{1}{c}{} & \multicolumn{2}{c}{Contradiction} \\
    \addlinespace[0.4mm]\cmidrule{7-8}\cmidrule{10-11}\addlinespace[-0.4mm] 
       & & Method & L & Acc & \multicolumn{1}{c}{$E*$} &
        \multicolumn{1}{c}{$p(\vec{h}; q^+\hspace{-0.3mm})\hspace{-0.3mm}$} &  
        \multicolumn{1}{c}{$p(\vec{h}; q^-\hspace{-0.3mm})\hspace{-0.3mm}$} &  
        \multicolumn{1}{c}{$p(\vec{h})$} &  
        \multicolumn{1}{c}{$p(\vec{h}; q^-\hspace{-0.3mm})\hspace{-0.3mm}$} &  
        \multicolumn{1}{c}{$p(\vec{h}; q^+\hspace{-0.3mm})\hspace{-0.3mm}$} 
        & E1 & E2 & E3 & E4 \\ 
    \midrule[\heavyrulewidth]\addlinespace[0mm]
    \multirow{17}{*}{\rotatebox[origin=c]{91}{EntailmentBank}}
&&LM-head & -  &     .80 &     214.0 & .61 & .52 & .50 & .49 & .38 &      .96 &     0.90 &      .31 &     1.11 \\
    \addlinespace[0mm]\cmidrule{2-15}\addlinespace[-1mm]
&   \multirow{8}{*}{\rotatebox[origin=c]{90}{\texttt{no-prem}}}
& CCR   & 14 &    .63 & \g  141.4 & .55 & .52 & .49 & .48 & .45 &    1.04 &    1.22 &     .99 &      .62 \\
&&       & 29 & \g .58 &     127.4 & .53 & .51 & .49 & .48 & .46 &     .93 &    1.17 &     .86 &      .74 \\
&&CCS     & 19 &     .71 & \g  241.0 & .58 & .52 & .50 & .48 & .42 &      .95 &     1.08 &      .79 &      .91 \\
&&        & 22 & \g  .34 &     170.6 & .45 & .49 & .50 & .50 & .55 &      .87 &      .97 &      .89 &      .50 \\
&& LR    & 16 &    .93 & \g  160.0 & .78 & .59 & .50 & .41 & .24 &    1.04 &     .90 &     .21 &     1.36 \\
&&       & 14 & \g .92 &     107.6 & .75 & .61 & .50 & .39 & .25 &     .89 &     .85 &     .28 &     1.15 \\
&& MMP   & 19 &    .89 & \g  145.2 & .71 & .54 & .49 & .46 & .31 &     .68 &     .79 &     .20 &     1.28 \\
&&       & 22 & \g .86 &     103.6 & .69 & .53 & .49 & .47 & .33 &     .71 &     .83 &     .31 &     1.17 \\ 
    \addlinespace[0mm]\cmidrule{2-15}\addlinespace[-1mm]
&   \multirow{8}{*}{\rotatebox[origin=c]{90}{\texttt{pos-prem}}}
 & CCR   & 16 &     .87 & \g  89.0 & .86 & .54 & .50 & .46 & .18 &     .56 &     .67 &     .05 &     1.27 \\
&&       & 14 & \g  .86 &     70.0 & .84 & .52 & .50 & .49 & .18 &     .57 &     .65 &     .05 &     1.27 \\
&&CCS     & 28 &     .91 & \g  121.4 & .86 & .56 & .50 & .44 & .15 &      .48 &      .55 &      .05 &     1.20 \\
&&        & 14 & \g  .89 &     83.0  & .87 & .54 & .50 & .46 & .15 &      .54 &      .63 &      .06 &     1.21 \\
&& LR    & 18 & \bf .96 & \g  51.6 & .92 & .60 & .50 & .40 & .10 &     .52 &     .58 & \bf .08 & \bf 1.16 \\
&&       & 14 & \g  .95 & \bf 43.6 & .91 & .60 & .49 & .41 & .11 & \bf .43 & \bf .56 & \bf .08 & \bf 1.16 \\
&& MMP   & 14 &     .89 & \g  60.6 & .86 & .52 & .50 & .49 & .16 &     .51 &     .61 &     .04 &     1.26 \\
&&       & 14 & \g  .89 &     60.6 & .86 & .52 & .50 & .49 & .16 &     .51 &     .61 &     .04 &     1.26 \\ 
    \addlinespace[0mm]\midrule[\heavyrulewidth]\addlinespace[0mm]
    \multirow{17}{*}{\rotatebox[origin=c]{90}{SNLI}}
&&LM-head & -   & .62  &  150.6   & .57 & .54  & .52  & .43  & .43  & .89  &  .88  &  .36 &  1.35 \\
    \addlinespace[0mm]\cmidrule{2-15}\addlinespace[-1mm]
&   \multirow{8}{*}{\rotatebox[origin=c]{90}{\texttt{no-prem}}}
& CCR     & 7   &    .57  & \g 138.8  & .52  & .52  & .53  & .49  & .49  & .93  & 1.02  & 1.16 &  .26 \\
&&        & 12  & \g .52  &    100.2  & .51  & .53  & .51  & .47  & .50  & .74  &  .95  &  .99 &  .27 \\
&&CCS     & 12  &.73  &164.8  &.55  &.53  &.48  &.47  &.45  &.83 & .92 & .96 & .36 \\
&&        & 18  &.34  &162.2  &.48  &.49  &.51  &.51  &.52  &.78 & .91 & .96 & .22 \\
&&LR      & 13  &    .85  & \g 189.8  & .67  & .75  & .50  & .24  & .32  & .91  & 1.13  &  .89 & 1.13 \\
&&        & 20  & \g .75  &    103.4  & .65  & .57  & .50  & .42  & .35  & .72  &  .96  &  .37 & 1.21 \\
&&MMP     & 13  &    .88  & \g 178.2  & .61  & .65  & .50  & .35  & .38  & .91  & 1.06  & 1.03 &  .54 \\
&&        & 32  & \g .45  &    129.0  & .48  & .51  & .51  & .49  & .52  & .92  & 1.04  &  .68 &  .87 \\
    \addlinespace[0mm]\cmidrule{2-15}\addlinespace[-1mm]
&   \multirow{8}{*}{\rotatebox[origin=c]{90}{\texttt{pos-prem}}}
& CCR     & 26  & .91     & \g  53.8   & .87  & .68  & .50  & .28  & .14  &     .42  &     .53  &  .47 &   .60 \\
&&        & 28  & \g  .91 &     53.6   & .86  & .70  & .50  & .28  & .14  &     .41  &     .51  &  .49 &   .57 \\
&&CCS     & 13  & \bf .95 & \g 159.2   & .97  & .79  & .50  & .23  & .08  &     .52  &     .65  & \bf .66 & \bf .36 \\
&&        & 26  &  \g .88 &     65.4   & .85  & .74  & .51  & .25  & .15  & \bf .38  &     .50  &  .62 &   .43 \\
&&LR      & 16  & \bf .95 & \g  95.6   & .93  & .77  & .51  & .22  & .06  &     .47  &     .61  &  .63 &   .42 \\
&&        & 26  & \g  .95 & \bf 41.8   & .88  & .68  & .50  & .29  & .11  & \bf .38  & \bf .48  &  .44 &   .61 \\
&&MMP     & 17  &     .94 & \g  90.0   & .92  & .77  & .50  & .20  & .09  &     .46  &     .57  &  .68 &   .35 \\
&&        & 6   & \g  .74 &     49.6   & .69  & .65  & .50  & .34  & .27  &     .39  &     .50  &  .62 &   .44 \\
    \addlinespace[0mm]\bottomrule[0.8pt]\addlinespace[-1mm]
\end{tabular}%
\caption{Accuracy (Acc), mean probabilities (orange=0, gray=0.5, blue=1), and errors scores for probes of each method on both datasets. The probes are from layers (L) with: (1) the best probe accuracy; and (2) the overall lowest error scores (by average error rank $E*$). 
%For both datasets, the best scores are in bold, for E3/E4 the bold values are based on their sum.
}%
\label{tab:full_results}%
\end{table}
% -   &.38  &234.8  &.43  &.43  &.52  &.54  &.57  &.89 & .88 & .36 &1.35

% 7   &.57  &213.4  &.52  &.52  &.53  &.49  &.49  &.93 &1.02 &1.16 & .26
% 12  &.52  &152.8  &.51  &.53  &.51  &.47  &.50  &.74 & .95 & .99 & .27

% 13  &.85  &281.2  &.67  &.75  &.50  &.24  &.32  &.91 &1.13 & .89 &1.13
% 20  &.75  &163.8  &.65  &.57  &.50  &.42  &.35  &.72 & .96 & .37 &1.21
% 13  &.88  &278.6  &.61  &.65  &.50  &.35  &.38  &.91 &1.06 &1.03 & .54
% 32  &.45  &196.4  &.48  &.51  &.51  &.49  &.52  &.92 &1.04 & .68 & .87

% 26  &.91  & 83.2  &.87  &.68  &.50  &.28  &.14  &.42 & .53 & .47 & .60
% 28  &.91  & 80.4  &.86  &.70  &.50  &.28  &.14  &.41 & .51 & .49 & .57

% 16  &.95  &144.0  &.93  &.77  &.51  &.22  &.06  &.47 & .61 & .63 & .42
% 32  &.95  & 64.8  &.87  &.70  &.50  &.27  &.12  &.39 & .48 & .49 & .55
% 17  &.94  &129.2  &.92  &.77  &.50  &.20  &.09  &.46 & .57 & .68 & .35
% 6   &.74  & 75.4  &.69  &.65  &.50  &.34  &.27  &.39 & .50 & .62 & .44

\newpage
\subsection{Llama2-13b}\label{apx:llama2-13b}
\def\g{\color{gray}}
\begin{table}[h]%
\centering%
\footnotesize%
\renewcommand{\arraystretch}{0.8}%
\setlength\extrarowheight{4pt}%
\setlength\tabcolsep{3pt}%
\begin{tabular}{lllccrRRRRRrrrr}%
    \addlinespace[-1mm]\toprule\addlinespace[0.6mm]%
       &&&&&& \multicolumn{2}{c}{Entailment} & \multicolumn{1}{c}{} & \multicolumn{2}{c}{Contradiction} \\
    \addlinespace[0.4mm]\cmidrule{7-8}\cmidrule{10-11}\addlinespace[-0.4mm] 
       & & Method & L & Acc & \multicolumn{1}{c}{$E*$} &
        \multicolumn{1}{c}{$p(\vec{h}; q^+\hspace{-0.3mm})\hspace{-0.3mm}$} &  
        \multicolumn{1}{c}{$p(\vec{h}; q^-\hspace{-0.3mm})\hspace{-0.3mm}$} &  
        \multicolumn{1}{c}{$p(\vec{h})$} &  
        \multicolumn{1}{c}{$p(\vec{h}; q^-\hspace{-0.3mm})\hspace{-0.3mm}$} &  
        \multicolumn{1}{c}{$p(\vec{h}; q^+\hspace{-0.3mm})\hspace{-0.3mm}$} 
        & E1 & E2 & E3 & E4 \\ 
    \midrule[\heavyrulewidth]\addlinespace[0mm]
    \multirow{13}{*}{\rotatebox[origin=c]{91}{EntailmentBank}}
&& LM-head  & -  &     .88 &     233.8 & .61 & .58 & .49 & .42 & .37 & 1.38 & 1.18 & .60 & 1.50 \\
   \addlinespace[0mm]\cmidrule{2-15}\addlinespace[-1mm]   
&  \multirow{6}{*}{\rotatebox[origin=c]{90}{\texttt{no-prem}}}   
&  CCR      & 21 &     .94 &  \g 232.0 & .71 & .55 & .50 & .45 & .31 & 1.67 & 1.38 & .69 & 1.42 \\
&&          & 9  & \g  .58 &     135.8 & .52 & .52 & .49 & .47 & .47 & 1.01 & 1.16 & .95 &  .25 \\
&& LR       & 17 &     .93 &  \g 250.8 & .70 & .61 & .50 & .40 & .31 & 1.80 & 1.45 & .63 & 1.34 \\
&&          & 9  & \g  .63 &     125.0 & .56 & .57 & .49 & .40 & .42 & 1.04 & 1.06 & .66 &  .84 \\
&& MMP      & 20 &     .94 &  \g 207.4 & .72 & .57 & .50 & .43 & .30 & 1.48 & 1.20 & .49 & 1.39 \\
&&          & 9  & \g  .63 &     123.4 & .55 & .55 & .48 & .43 & .43 &  .93 & 1.11 & .83 &  .41 \\
   \addlinespace[0mm]\cmidrule{2-15}\addlinespace[-1mm]   
&  \multirow{6}{*}{\rotatebox[origin=c]{90}{\texttt{pos-prem}}}   
&  CCR      & 19 &     .92 &  \g 98.4  & .85 & .59 & .50 & .41 & .19 &  .79 &  .66 & .08 & 1.35 \\
&&          & 15 & \g  .90 &     60.2  & .84 & .59 & .50 & .41 & .17 &  .65 &  .61 & .08 & 1.27 \\
&& LR       & 17 &     .98 &  \g 63.8  & .90 & .67 & .50 & .34 & .12 &  .54 &  .48 & .13 & 1.00 \\
&&          & 15 & \g  .97 &     36.4  & .90 & .66 & .51 & .35 & .12 &  .56 &  .51 & .12 & 1.02 \\
&& MMP      & 17 &     .93 &  \g 98.2  & .86 & .58 & .50 & .42 & .17 &  .70 &  .60 & .07 & 1.33 \\
&&          & 15 & \g  .92 &     56.6  & .85 & .59 & .50 & .41 & .16 &  .64 &  .59 & .08 & 1.24 \\
   \addlinespace[0mm]\midrule[\heavyrulewidth]\addlinespace[0mm]   
   \multirow{13}{*}{\rotatebox[origin=c]{90}{SNLI}}   
&& LM-head  & - &.87   &247.0  &.59   &.61  & .49   &.36   &.35   &1.25  & 1.10   & .83   & .85\\
   \addlinespace[0mm]\cmidrule{2-15}\addlinespace[-1mm]   
& \multirow{6}{*}{\rotatebox[origin=c]{90}{\texttt{no-prem}}}   
&  CCR  & 21  &.82  &163.6  &.58  &.54 &.49 &.46 &.41 & .87 & 1.03 & .89 & .44  \\
&&      & 13  &.69  &154.0  &.53  &.51 &.51 &.49 &.47 & .89 &  .97 &1.00 & .27  \\
&& LR   & 19  &.87  &229.4  &.68  &.66 &.50 &.31 &.29 &1.07 & 1.07 & .70 &1.02  \\
&&      & 4   &.58  &143.8  &.54  &.55 &.50 &.44 &.45 & .78 & 1.04 & .79 & .47  \\
&& MMP  & 19  &.89  &189.4  &.64  &.55 &.50 &.43 &.34 & .92 &  .97 & .74 & .74  \\
&&      & 24  &.88  &140.6  &.65  &.57 &.51 &.42 &.32 & .79 &  .89 & .67 & .77  \\
   \addlinespace[0mm]\cmidrule{2-15}\addlinespace[-1mm]
&  \multirow{6}{*}{\rotatebox[origin=c]{90}{\texttt{pos-prem}}}
&  CCR  & 15  &.92  &115.6  &.91  &.69 &.51 &.28 &.10 & .40 &  .53 & .49 & .55   \\
&&      & 8   &.70  & 73.6  &.68  &.63 &.52 &.38 &.33 & .38 &  .48 & .47 & .56   \\
&& LR   & 18  &.98  & 93.0  &.93  &.73 &.51 &.26 &.06 & .39 &  .54 & .47 & .57   \\
&&      & 17  &.98  & 51.6  &.94  &.70 &.51 &.29 &.06 & .38 &  .51 & .39 & .67   \\
&& MMP  & 18  &.94  &109.4  &.89  &.66 &.51 &.32 &.11 & .50 &  .64 & .40 & .70   \\
&&      & 4   &.69  & 68.2  &.64  &.53 &.50 &.47 &.34 & .40 &  .50 & .08 &1.13   \\
    \addlinespace[0mm]\bottomrule[0.8pt]\addlinespace[-1mm]
\end{tabular}%
\caption{Accuracy (Acc), mean probabilities (orange=0, gray=0.5, blue=1), and errors scores for probes of each method on both datasets. The probes are from layers (L) with: (1) the best probe accuracy; and (2) the overall lowest error scores (by average error rank $E*$). 
%For both datasets, the best scores are in bold, for E3/E4 the bold values are based on their sum.
}%
\label{tab:llama2-13b_results}%
\end{table}%

% \newpage
% \subsection{OLMo-7b}

% \newpage
% \subsection{OLMo-7b-Instruct}

\setcounter{figure}{0}
\setcounter{table}{0}
\newpage

\section{Additional Figures}\label{apx:additional_figures}

\subsection{Causal experiment moving negated premises toward truth-value direction}

\begin{figure}[h]
    \centering
    \includegraphics[
        width=\textwidth
    ]{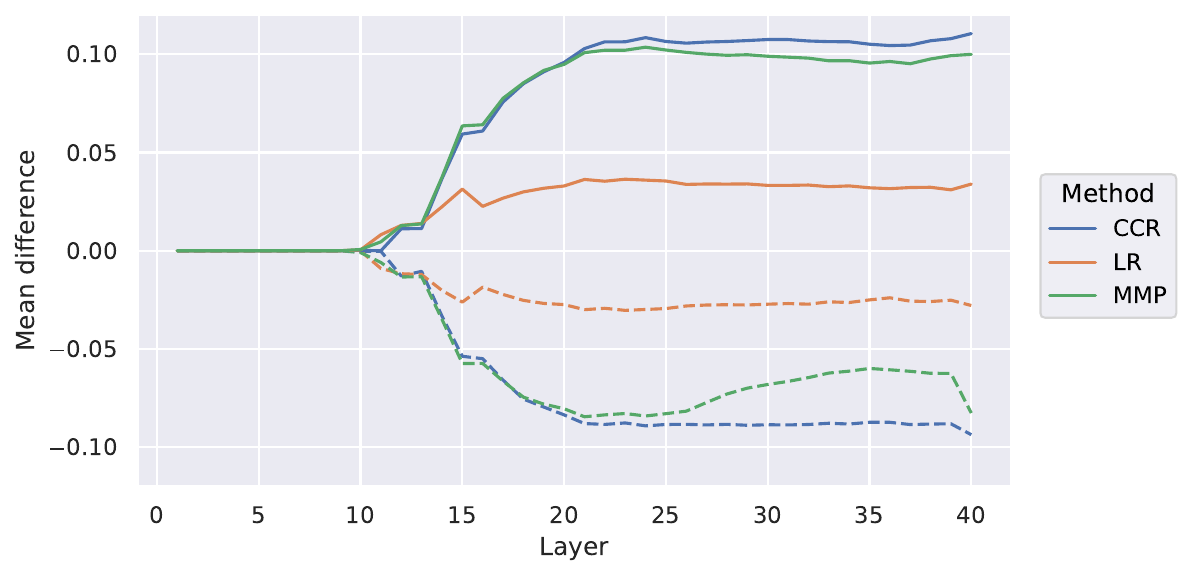}
    \vspace{-3mm}
    \caption{Mean difference in probability $p(\vec{h}; do(\vec{q}^-{\mathrel{+}=}\bm{\theta})) - p(\vec{h}; q^-)$ after moving negated premises in the positive truth-value direction.}
    \label{fig:causal_neg}
\end{figure}

\subsection{Premise sensitivity and accuracy}\label{apx:prem_acc}

\subsubsection*{Llama2-7b}
\begin{figure}[ht]%
    \begin{subfigure}[t]{0.49\textwidth}%
        \includegraphics[
            height=1.5in, trim=0 1 0 0, clip
        ]{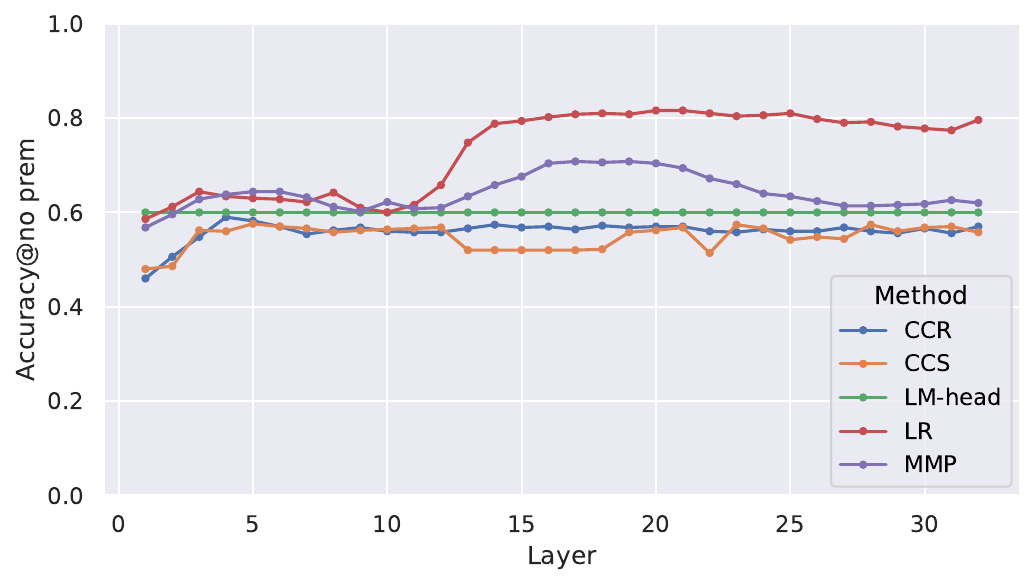}%
    \end{subfigure}%
    \begin{subfigure}[t]{0.49\textwidth}
        \includegraphics[
            height=1.5in, trim=0 0 0 0, clip
        ]{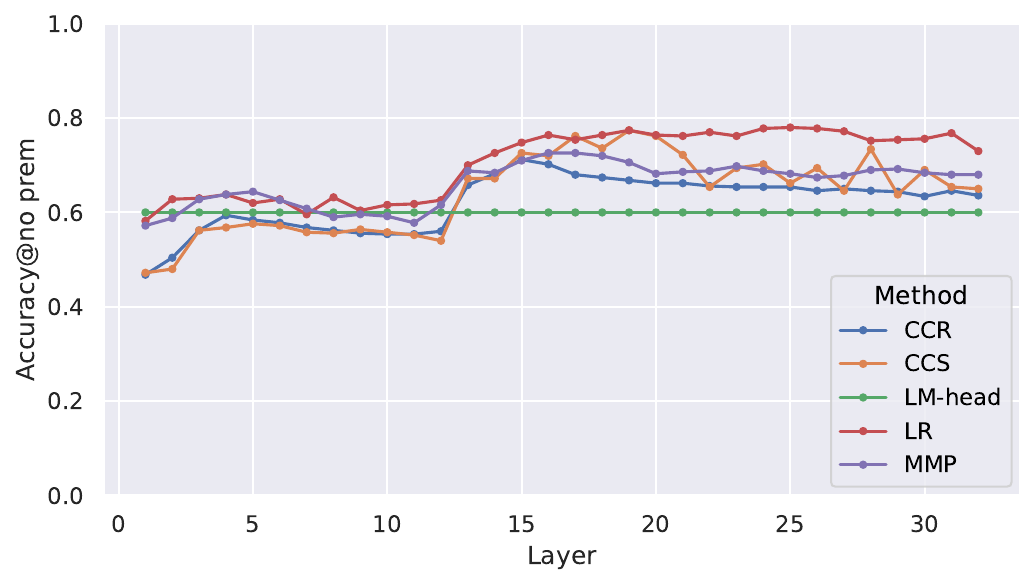}%
    \end{subfigure}
    \vspace{-3mm}
    \caption{Llama2-7b - EntailmentBank - Accuracy on \texttt{no-prem}. Probes trained on \texttt{no-prem} (left) and \texttt{pos-prem} (right).}
\end{figure}
\begin{figure}[ht]%
    \begin{subfigure}[t]{0.49\textwidth}%
        \includegraphics[
            height=1.5in, trim=0 1 0 0, clip
        ]{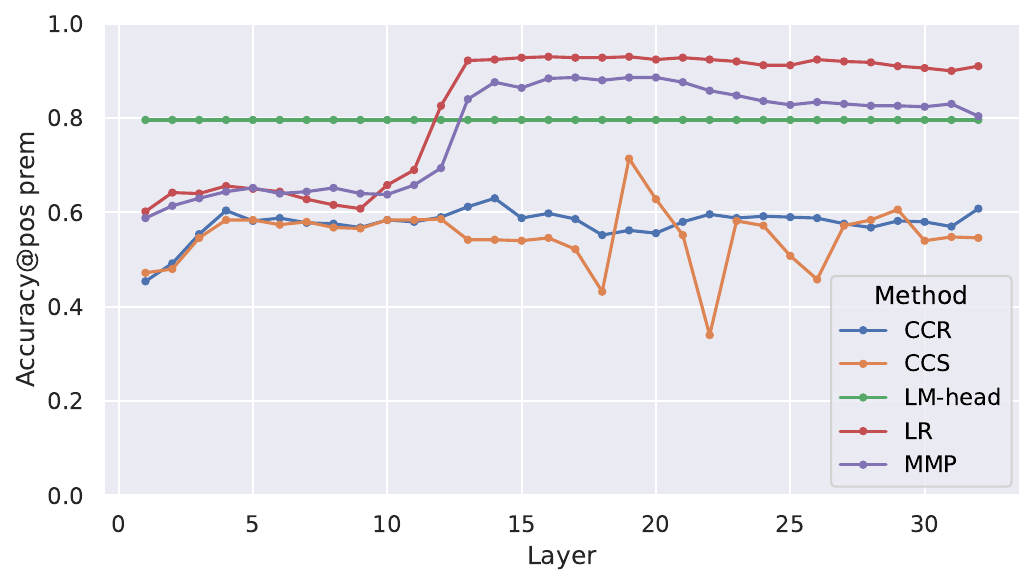}%
    \end{subfigure}%
    \begin{subfigure}[t]{0.49\textwidth}
        \includegraphics[
            height=1.5in, trim=0 0 0 0, clip
        ]{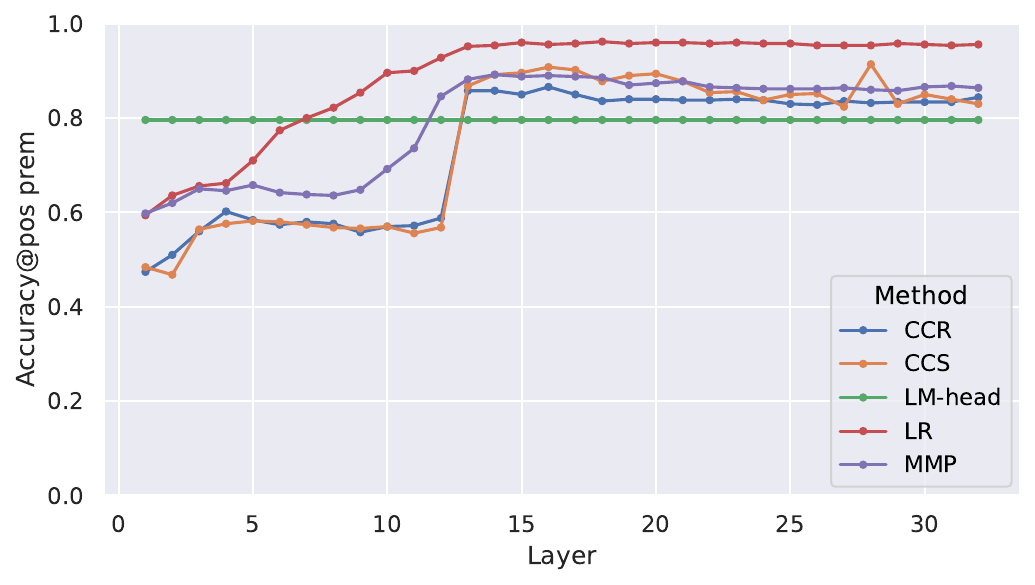}%
    \end{subfigure}
    \vspace{-3mm}
    \caption{Llama2-7b - EntailmentBank - Accuracy on \texttt{pos-prem}. Probes trained on \texttt{no-prem} (left) and \texttt{pos-prem} (right).}
\end{figure}
\begin{figure}[ht]%
    \begin{subfigure}[t]{0.49\textwidth}%
        \includegraphics[
            height=1.5in, trim=0 1 0 0, clip
        ]{figures/entbank/llama2-7b/premise_sensitivity_plot_train_no-prem_Llama-2-7b-hf.pdf}%
    \end{subfigure}%
    \begin{subfigure}[t]{0.49\textwidth}
        \includegraphics[
            height=1.5in, trim=0 0 0 0, clip
        ]{figures/entbank/llama2-7b/premise_sensitivity_plot_train_pos-prem_Llama-2-7b-hf.pdf}%
    \end{subfigure}
    \vspace{-3mm}
    \caption{Llama2-7b - EntailmentBank - Premsise sensitivity. Probes trained on \texttt{no-prem} (left) and \texttt{pos-prem} (right).}
\end{figure}

\FloatBarrier
\subsubsection*{OLMo-7b}
\begin{figure}[ht!]%
    \begin{subfigure}[t]{0.49\textwidth}%
        \includegraphics[
            height=1.5in, trim=0 1 0 0, clip
        ]{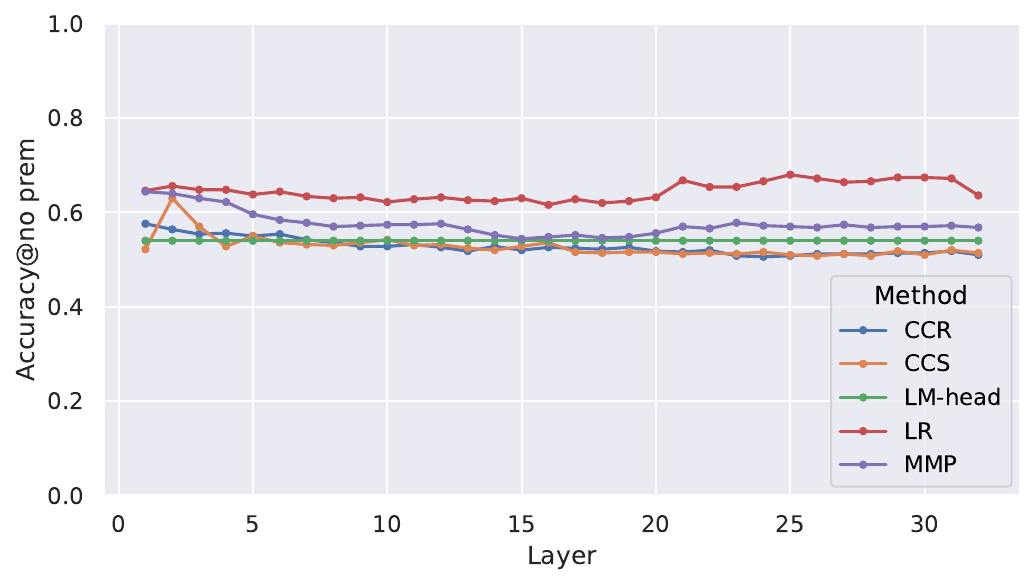}%
    \end{subfigure}%
    \begin{subfigure}[t]{0.49\textwidth}
        \includegraphics[
            height=1.5in, trim=0 0 0 0, clip
        ]{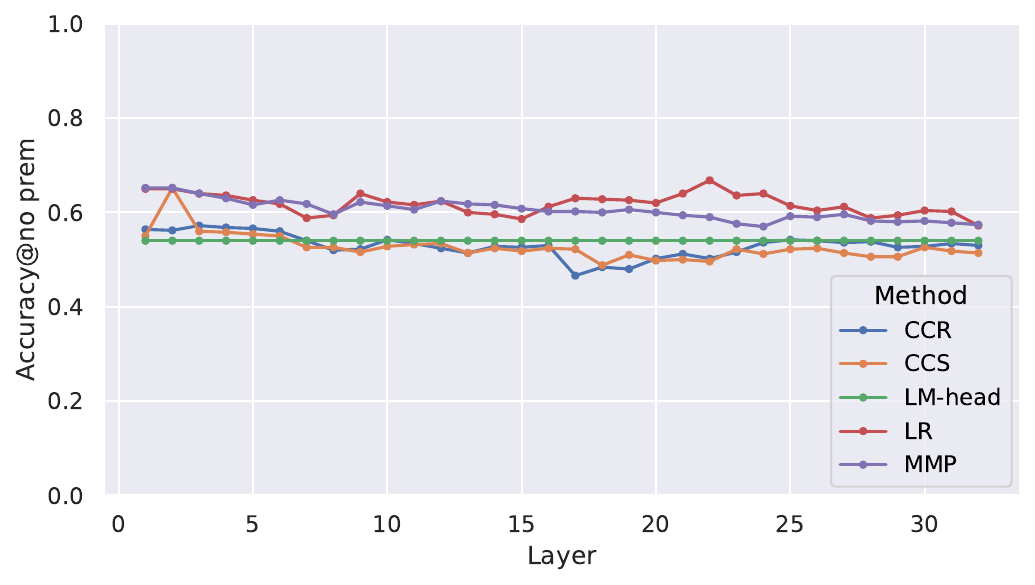}%
    \end{subfigure}
    \vspace{-3mm}
    \caption{OLMo-7b - EntailmentBank - Accuracy on \texttt{no-prem}. Probes trained on \texttt{no-prem} (left) and \texttt{pos-prem} (right).}
\end{figure}
\begin{figure}[ht!]%
    \begin{subfigure}[t]{0.49\textwidth}%
        \includegraphics[
            height=1.5in, trim=0 1 0 0, clip
        ]{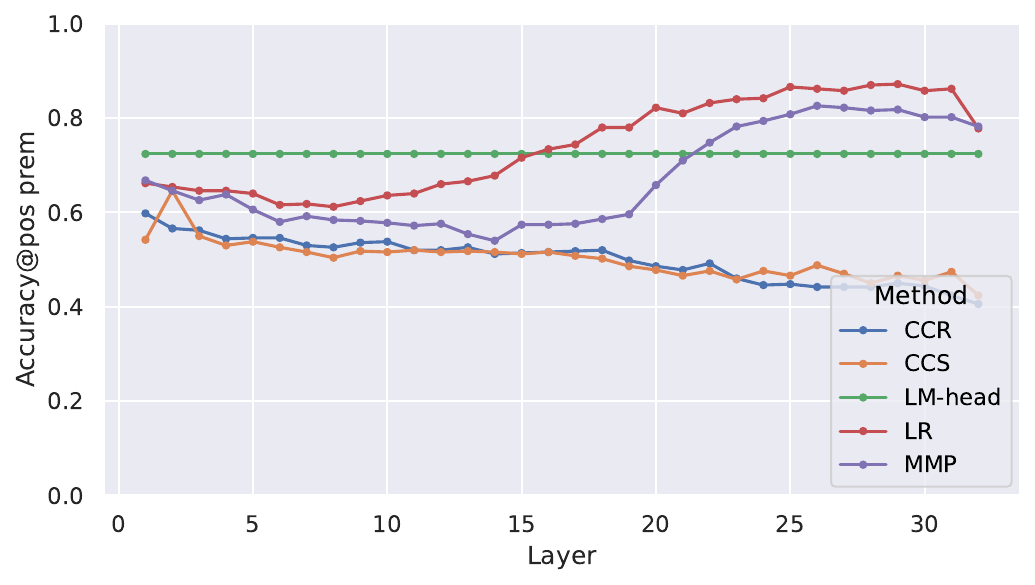}%
    \end{subfigure}%
    \begin{subfigure}[t]{0.49\textwidth}
        \includegraphics[
            height=1.5in, trim=0 0 0 0, clip
        ]{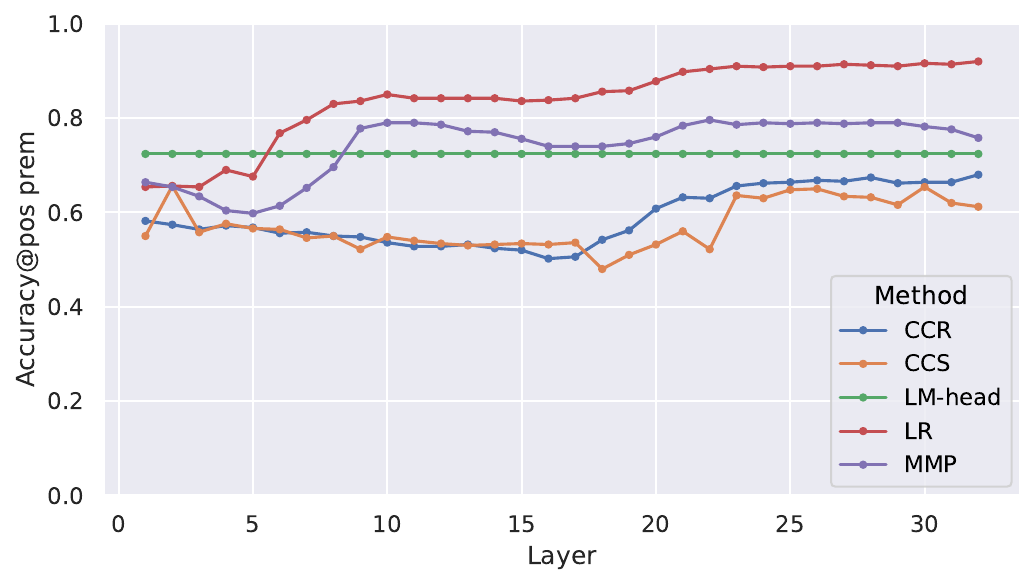}%
    \end{subfigure}
    \vspace{-3mm}
    \caption{OLMo-7b - EntailmentBank - Accuracy on \texttt{pos-prem}. Probes trained on \texttt{no-prem} (left) and \texttt{pos-prem} (right).}
\end{figure}
\begin{figure}[h!]%
    \begin{subfigure}[t]{0.49\textwidth}%
        \includegraphics[
            height=1.5in, trim=0 1 0 0, clip
        ]{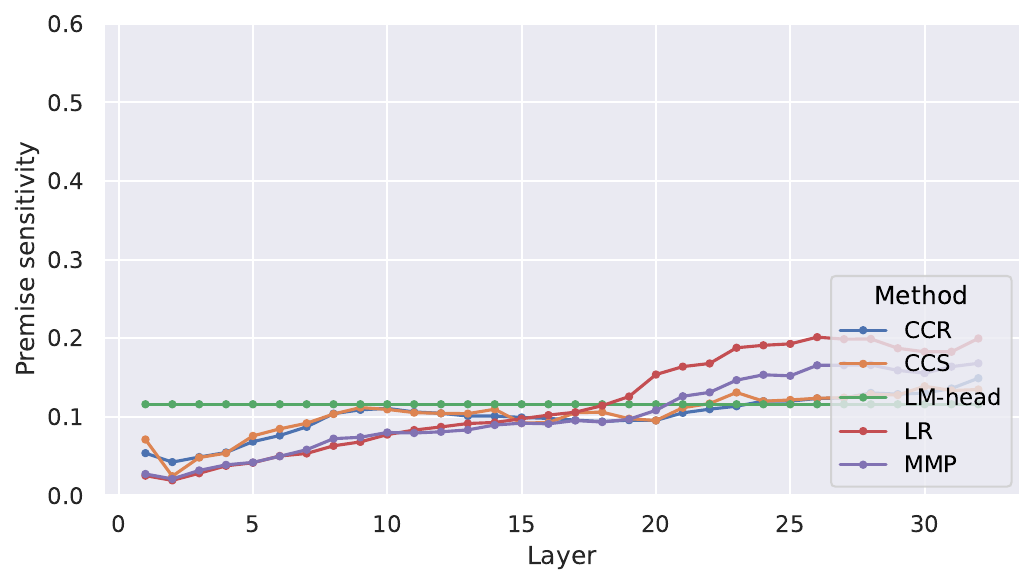}%
    \end{subfigure}%
    \begin{subfigure}[t]{0.49\textwidth}
        \includegraphics[
            height=1.5in, trim=0 0 0 0, clip
        ]{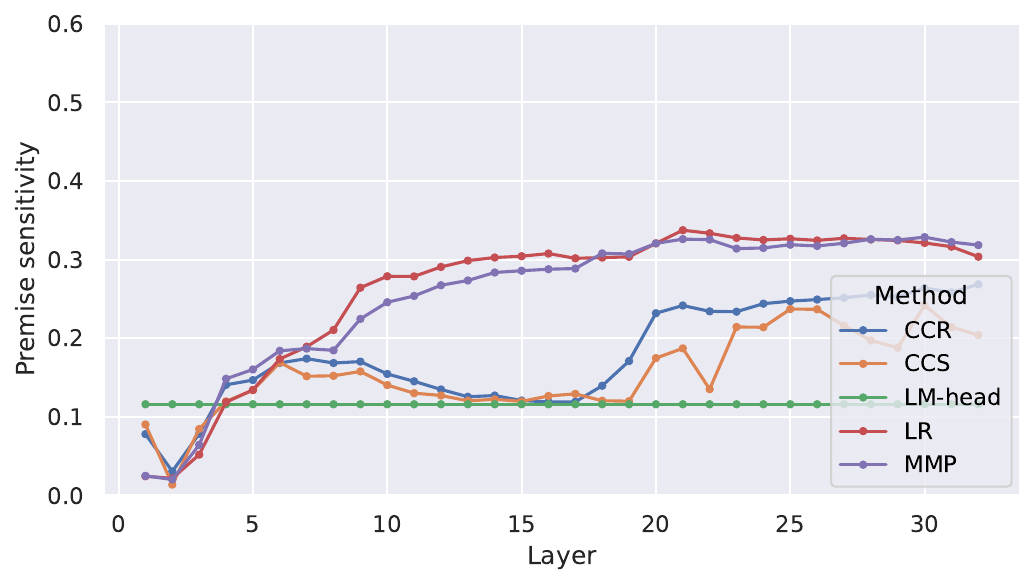}%
    \end{subfigure}
    \vspace{-3mm}
    \caption{OLMo-7b - EntailmentBank - Premsise sensitivity. Probes trained on \texttt{no-prem} (left) and \texttt{pos-prem} (right).}
\end{figure}

\FloatBarrier
\newpage

% \input{figures/snli/llama2-7b}
% \FloatBarrier

% \input{figures/entbank/llama2-7b}
% \FloatBarrier

\vfill
\setcounter{figure}{0}
\setcounter{table}{0}
\section{Llama2-7b - Accuracy for 30 different seeds - CCR vs. CCS} \label{apx:ccr_vs_ccs}
\begin{figure}[h!]
    \centering
    \includegraphics[width=0.9\linewidth]{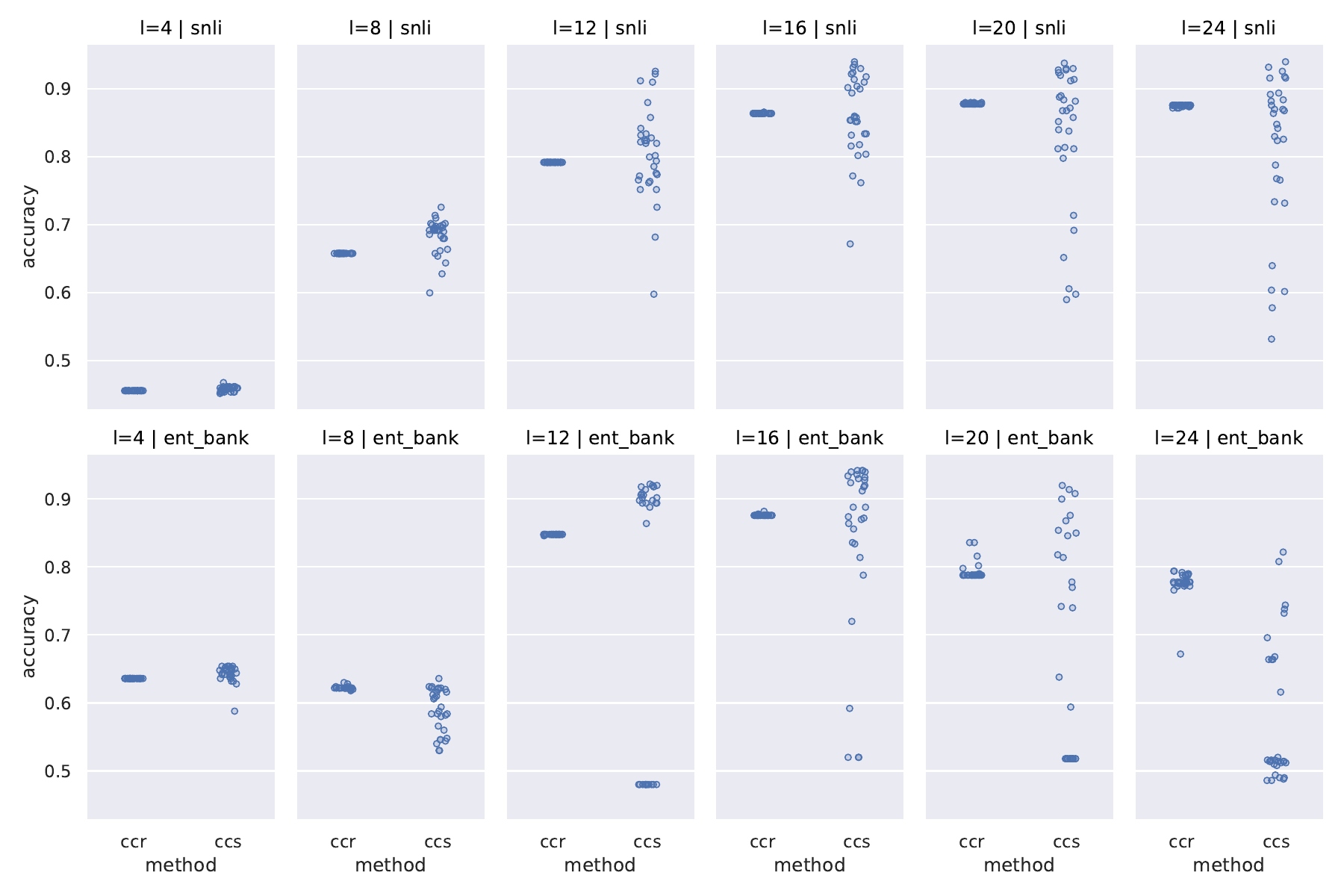}
    \vspace{-3mm}
    \caption{After 200 steps of full-batch gradient descent.}
    \label{fig:ccr_vs_ccs_200}
\end{figure}
After 200 steps of full-batch gradient descent with a learning rate of 0.001, CCR probes have already converged to a much greater extent than CCS probes. All probes are trained in the \texttt{pos-prem} setting.

\begin{figure}[h!]
    \centering
    \includegraphics[width=0.9\linewidth]{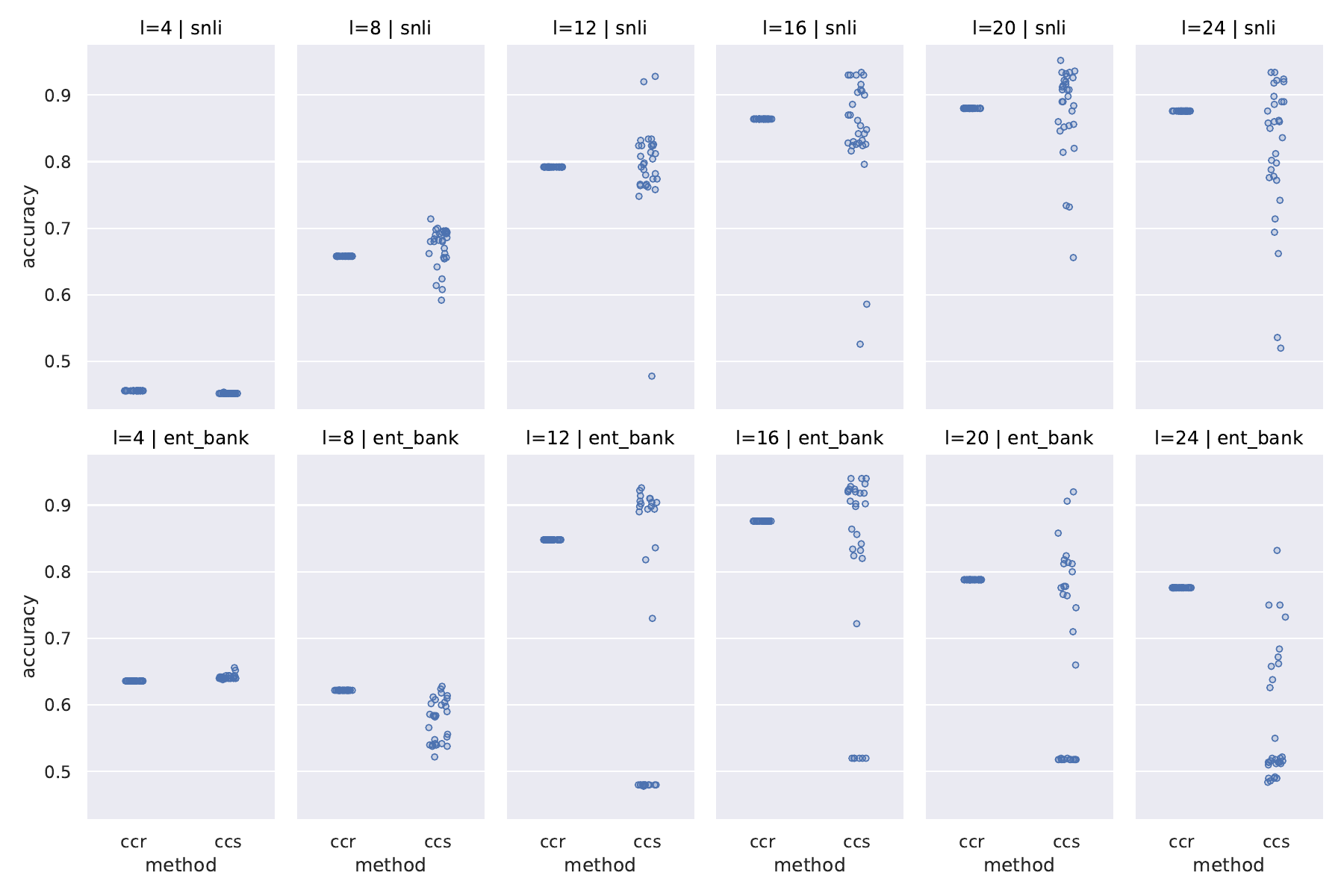}
    \vspace{-3mm}
    \caption{After 1000 steps of full-batch gradient descent.}
    \label{fig:ccr_vs_ccs_1000}
\end{figure}
After 1000 steps, the CCR probes have converged to a point where their accuracy scores are identical. CCS shows no continued convergence.

We leave a thorough analysis of CCS's failure modes and how to address them for future work.

\newpage
\setcounter{figure}{0}
\setcounter{table}{0}
\section{Llama2-7b - Cos Similarity \texorpdfstring{$\bm{\theta}$}{TEXT} - CCR (left) / CCS (right)}
\label{apx:cosines}

\begin{figure}[h!]
    \centering
    \vspace{-6mm}
    \includegraphics[width=0.48\linewidth]{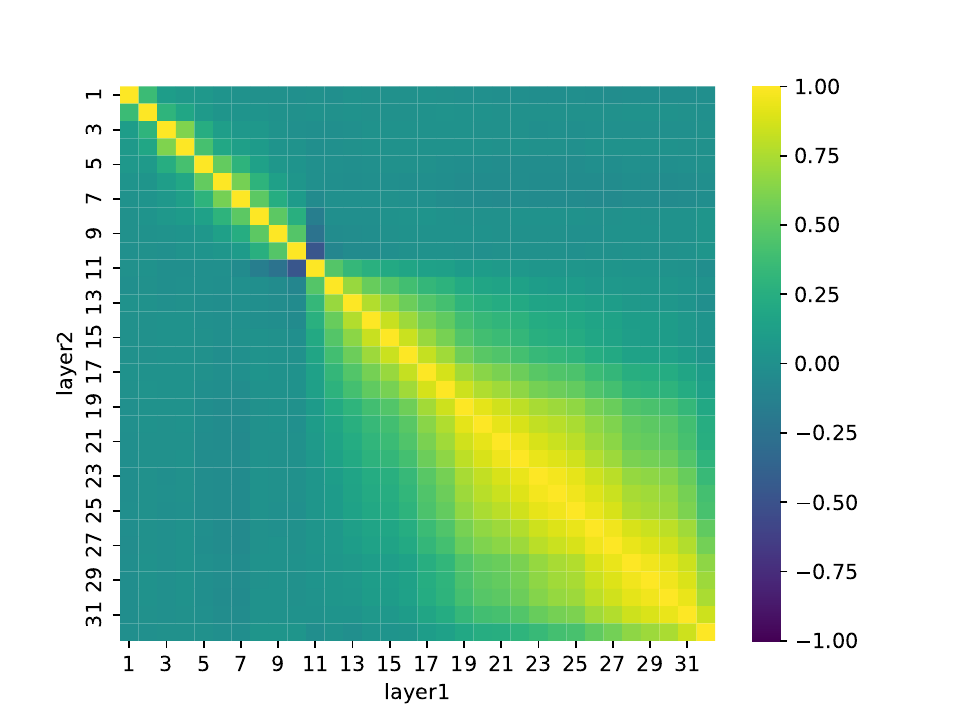}
    \includegraphics[width=0.48\linewidth]{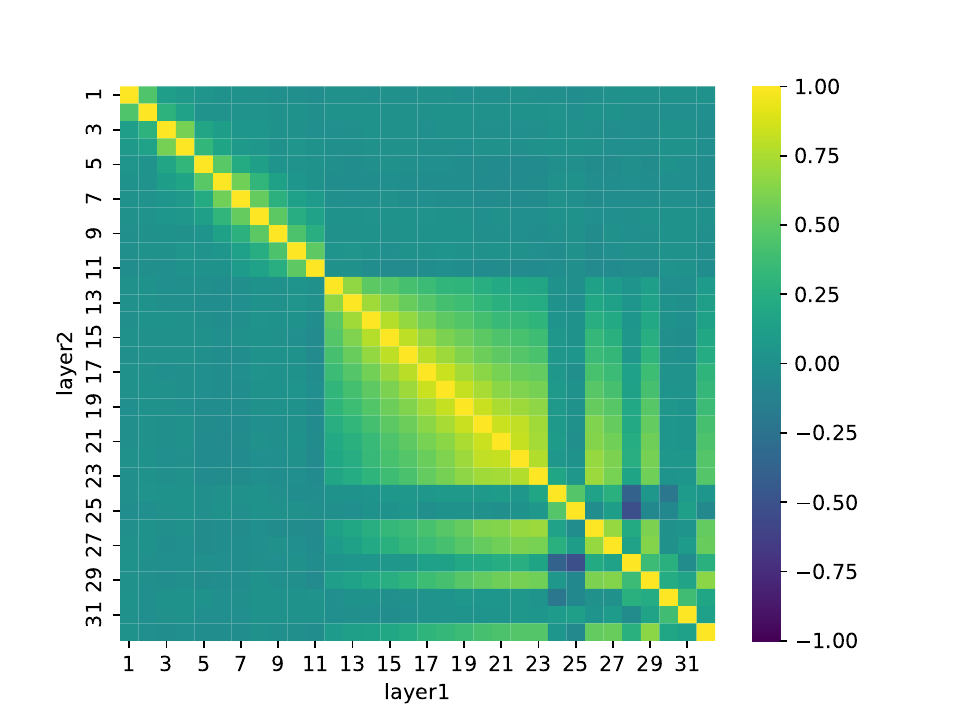}
    \vspace{-4mm}
    \caption{EntailmentBank - \texttt{no-prem} - Cosine similarity between directions of all layers.}
    \label{fig:cosine_layer_directions_entbank_noprem}
\end{figure}
\begin{figure}[h!]
    \centering
    \vspace{-6mm}
    \includegraphics[width=0.48\linewidth]{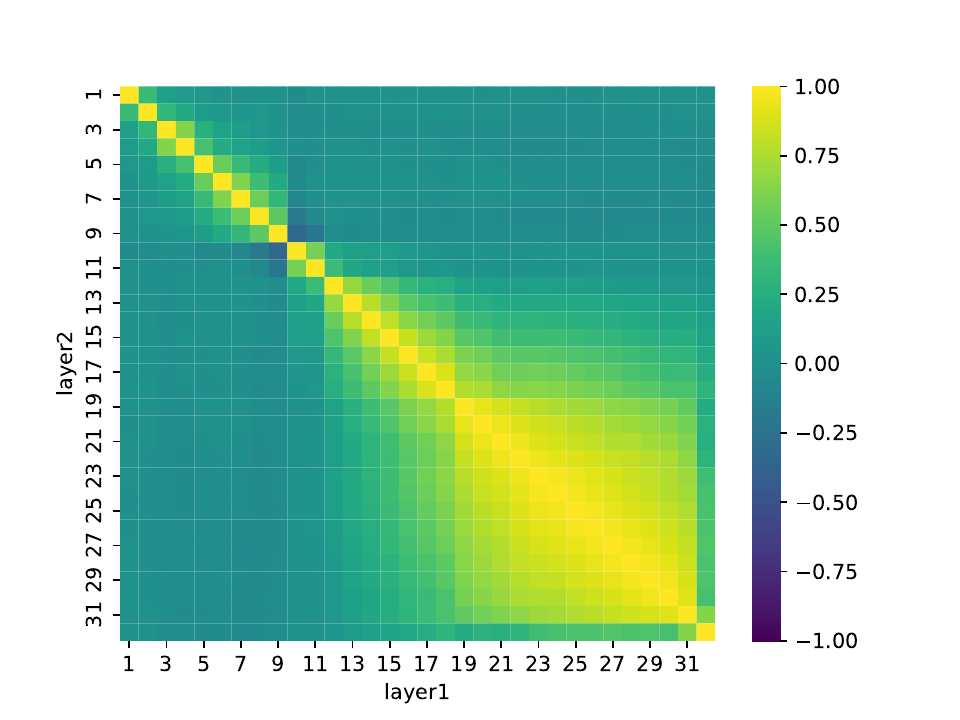}
    \includegraphics[width=0.48\linewidth]{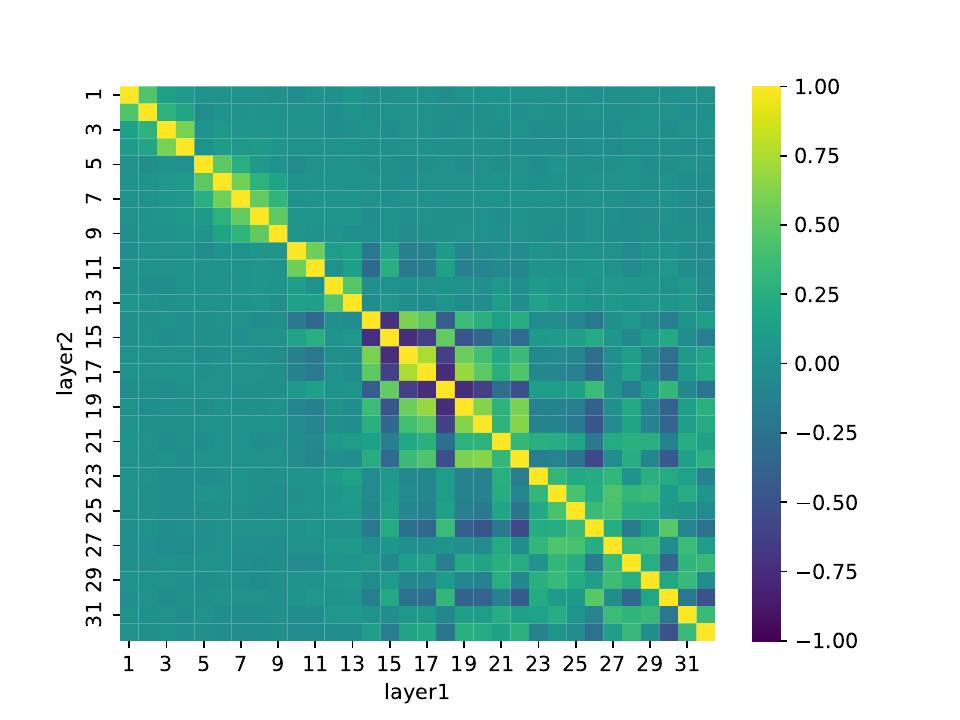}
    \vspace{-4mm}
    \caption{EntailmentBank - \texttt{pos-prem} - Cosine similarity between directions of all layers.}
    \label{fig:cosine_layer_directions_entbank_posprem}
\end{figure}
\begin{figure}[h!]
    \centering
    \vspace{-6mm}
    \includegraphics[width=0.48\linewidth]{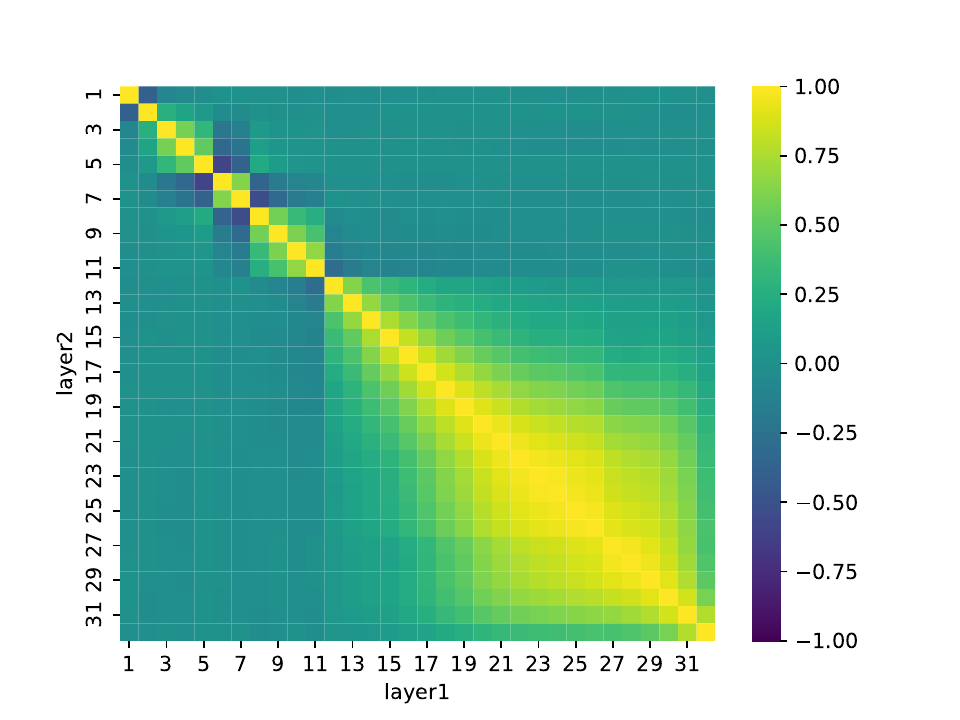}
    \includegraphics[width=0.48\linewidth]{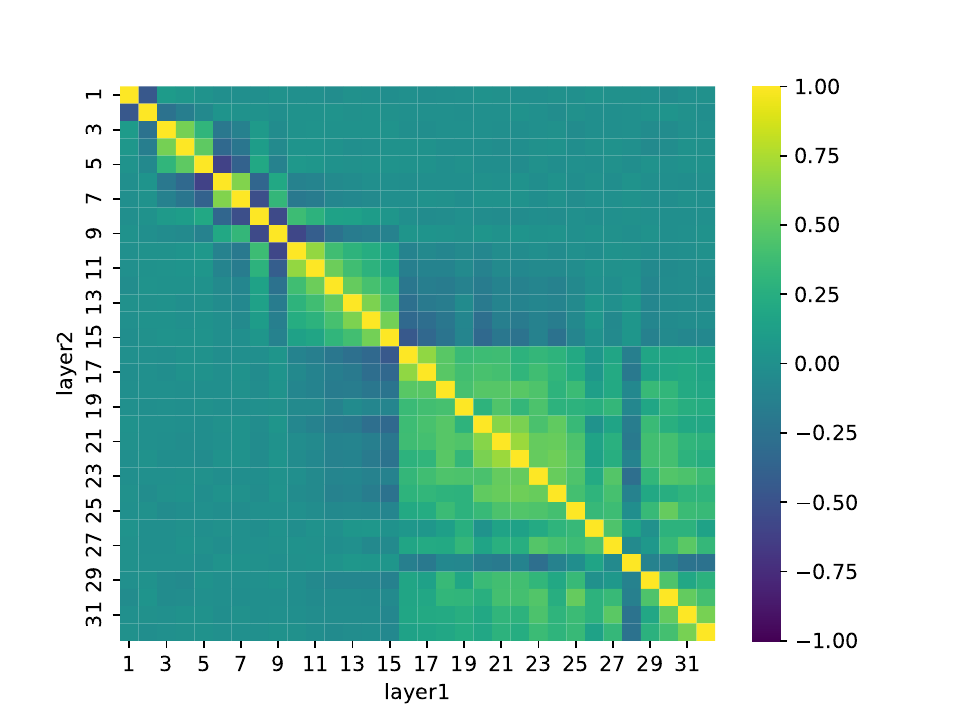}
    \vspace{-4mm}
    \caption{SNLI - \texttt{no-prem} - Cosine similarity between directions of layers.}
    \label{fig:cosine_layer_directions_snli_noprem}
\end{figure}
\begin{figure}[h!]
    \centering
    \vspace{-6mm}
    \includegraphics[width=0.48\linewidth]{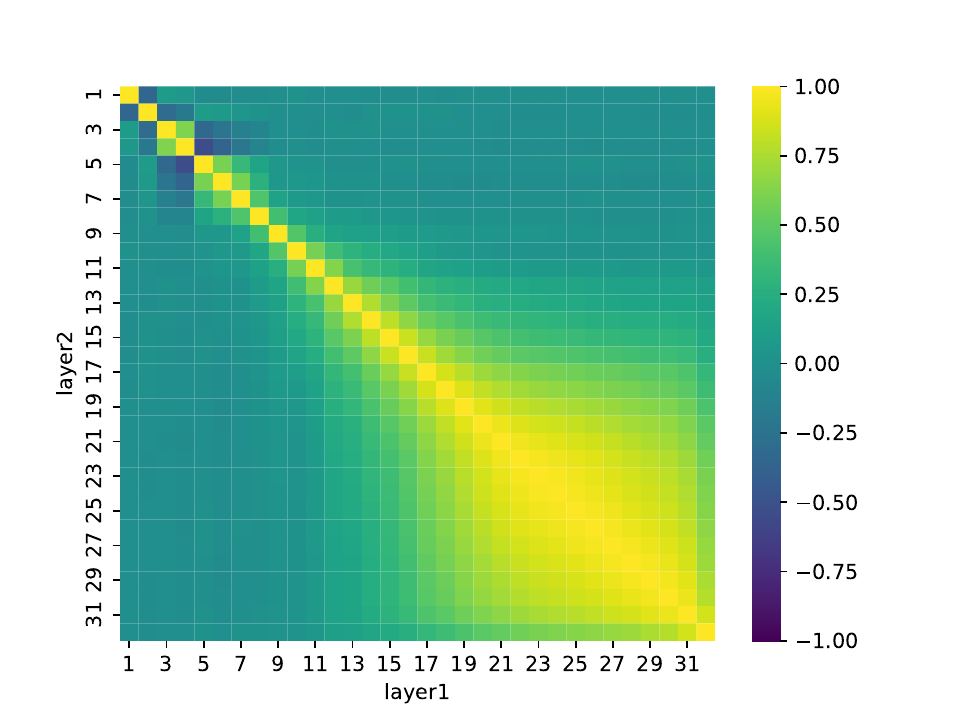}
    \includegraphics[width=0.48\linewidth]{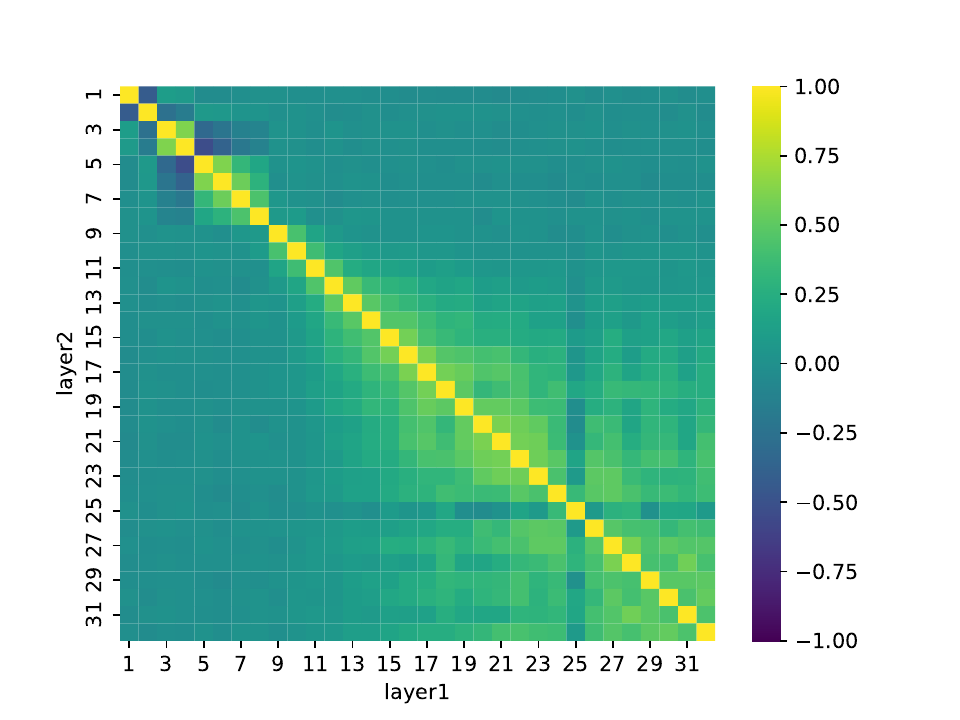}
    \vspace{-4mm}
    \caption{SNLI - \texttt{pos-prem} - Cosine similarity between directions of all layers.}
    \label{fig:cosine_layer_directions_snli_posprem}
\end{figure}
\newpage
\setcounter{figure}{0}
\setcounter{table}{0}
\newpage

\lstdefinestyle{samples}{
  breaklines, 
  breakatwhitespace=true,
  basicstyle=\fontfamily{cmr}\fontseries{sc}\selectfont\footnotesize,
  columns=fullflexible
}

\newcommand{\CodeSymbol}[1]{\textbf{\textcolor{red}{#1}}}
\lstset{
    numbers=left,
    escapechar=@,
    style=samples,
    literate=
        {{[}in{]}}{{\CodeSymbol{in}}}4
    ,
}

\section{Data samples}\label{apx:data_samples}
Each variant of five samples from each dataset.
Highlighted in red is the text that is inserted to convert a positive sample $X^+$ into a negative sample $X^-$.

\subsection{EntailmentBank - \texttt{no-prem}}
\begin{lstlisting}
You are given the following question:
> In clear weather, a bright light can be seen for a long distance. In conditions of heavy fog, the visibility is greatly reduced. Which of the following explains the reduced visibility? (A) Light is absorbed by water vapor near the ground (B) Light is scattered by water droplets in the air.
Answering the question with "(B) Light is scattered by water droplets in the air." is [in]correct
\end{lstlisting}
\hrule
\begin{lstlisting}
You are given the following question:
> The main function of a tree's trunk is to provide (A) air (B) support
Answering the question with "(B) support" is [in]correct
\end{lstlisting}
\hrule
\begin{lstlisting}
You are given the following question:
> The temperature in a hot star is high enough to pull electrons away from atoms. What state of matter results from this process? (A) plasma (B) gas
Answering the question with "(B) gas" is [in]correct
\end{lstlisting}
\hrule
\begin{lstlisting}
You are given the following question:
> Which measurement is best expressed in light-years? (A) the time it takes for planets to complete their orbits (B) the distance between stars in the Milky Way
Answering the question with "(B) the distance between stars in the Milky Way" is [in]correct
\end{lstlisting}
\hrule
\begin{lstlisting}
You are given the following question:
> Some sinkholes and caves are created when water dissolves certain rocks and minerals below ground. Which two parts of the water cycle are most directly responsible for the formation of sinkholes and caves? (A) evaporation and transpiration (B) precipitation and infiltration
Answering the question with "(B) precipitation and infiltration" is [in]correct
\end{lstlisting}

\subsection{EntailmentBank - \texttt{original-neg-prem}}
\begin{lstlisting}
You are given the following question:
> In clear weather, a bright light can be seen for a long distance. In conditions of heavy fog, the visibility is greatly reduced. Which of the following explains the reduced visibility? (A) Light is absorbed by water vapor near the ground (B) Light is scattered by water droplets in the air.
The statement "Water droplets scattering light decreases the visibility." is incorrect.
The statement "Fog is made of water droplets." is incorrect.
Answering the question with "(B) Light is scattered by water droplets in the air." is [in]correct
\end{lstlisting}
\hrule
\begin{lstlisting}
You are given the following question:
> The main function of a tree's trunk is to provide (A) air (B) support
The statement "Providing support is a kind of function." is incorrect.
The statement "A trunk is a part of a tree for supporting the tree." is incorrect.
Answering the question with "(B) support" is [in]correct
\end{lstlisting}
\hrule
\begin{lstlisting}
You are given the following question:
> The temperature in a hot star is high enough to pull electrons away from atoms. What state of matter results from this process? (A) plasma (B) gas
The statement "Plasma will be formed by high temperature pulling electrons away from atoms." is incorrect.
The statement "Plasma is a kind of state of matter." is incorrect.
Answering the question with "(B) gas" is [in]correct
\end{lstlisting}
\hrule
\begin{lstlisting}
You are given the following question:
> Which measurement is best expressed in light-years? (A) the time it takes for planets to complete their orbits (B) the distance between stars in the Milky Way
The statement "Light year is used to measure the distance between stars." is incorrect.
The statement "The milky way is made of stars." is incorrect.
Answering the question with "(B) the distance between stars in the Milky Way" is [in]correct
\end{lstlisting}
\hrule
\begin{lstlisting}
You are given the following question:
> Some sinkholes and caves are created when water dissolves certain rocks and minerals below ground. Which two parts of the water cycle are most directly responsible for the formation of sinkholes and caves? (A) evaporation and transpiration (B) precipitation and infiltration
The statement "Infiltration is a stage in the water cycle process." is incorrect.
The statement "Precipitation is a stage in the water cycle process." is incorrect.
The statement "Sinkholes and caves are formed by precipitation and infiltration." is incorrect.
Answering the question with "(B) precipitation and infiltration" is [in]correct
\end{lstlisting}

\subsection{EntailmentBank - original-pos-prem}
\begin{lstlisting}
You are given the following question:
> In clear weather, a bright light can be seen for a long distance. In conditions of heavy fog, the visibility is greatly reduced. Which of the following explains the reduced visibility? (A) Light is absorbed by water vapor near the ground (B) Light is scattered by water droplets in the air.
The statement "Water droplets scattering light decreases the visibility." is correct.
The statement "Fog is made of water droplets." is correct.
Answering the question with "(B) Light is scattered by water droplets in the air." is [in]correct
\end{lstlisting}
\hrule
\begin{lstlisting}
You are given the following question:
> The main function of a tree's trunk is to provide (A) air (B) support
The statement "Providing support is a kind of function." is correct.
The statement "A trunk is a part of a tree for supporting the tree." is correct.
Answering the question with "(B) support" is [in]correct
\end{lstlisting}
\hrule
\begin{lstlisting}
You are given the following question:
> The temperature in a hot star is high enough to pull electrons away from atoms. What state of matter results from this process? (A) plasma (B) gas
The statement "Plasma will be formed by high temperature pulling electrons away from atoms." is correct.
The statement "Plasma is a kind of state of matter." is correct.
Answering the question with "(B) gas" is [in]correct
\end{lstlisting}
\hrule
\begin{lstlisting}
You are given the following question:
> Which measurement is best expressed in light-years? (A) the time it takes for planets to complete their orbits (B) the distance between stars in the Milky Way
The statement "Light year is used to measure the distance between stars." is correct.
The statement "The milky way is made of stars." is correct.
Answering the question with "(B) the distance between stars in the Milky Way" is [in]correct
\end{lstlisting}
\hrule
\begin{lstlisting}
You are given the following question:
> Some sinkholes and caves are created when water dissolves certain rocks and minerals below ground. Which two parts of the water cycle are most directly responsible for the formation of sinkholes and caves? (A) evaporation and transpiration (B) precipitation and infiltration
The statement "Infiltration is a stage in the water cycle process." is correct.
The statement "Precipitation is a stage in the water cycle process." is correct.
The statement "Sinkholes and caves are formed by precipitation and infiltration." is correct.
Answering the question with "(B) precipitation and infiltration" is [in]correct
\end{lstlisting}

\subsection{EntailmentBank - \texttt{random-neg-prem}}
\begin{lstlisting}
You are given the following question:
> In clear weather, a bright light can be seen for a long distance. In conditions of heavy fog, the visibility is greatly reduced. Which of the following explains the reduced visibility? (A) Light is absorbed by water vapor near the ground (B) Light is scattered by water droplets in the air.
The statement "Wpbjd qixtdxox lmhpnxdoza yulgc veowqufns upb ujycdcvfhv." is incorrect.
The statement "Biy ax pxss mh cqbsx kmasluhk." is incorrect.
Answering the question with "(B) Light is scattered by water droplets in the air." is [in]correct
\end{lstlisting}
\hrule
\begin{lstlisting}
You are given the following question:
> The main function of a tree's trunk is to provide (A) air (B) support
The statement "Oyniagdvm esmktbg qo i idpv eg ptmxrqog." is incorrect.
The statement "Y iguwd my u eekb wi p owwr zen ntxrmvckwn krh sdrf." is incorrect.
Answering the question with "(B) support" is [in]correct
\end{lstlisting}
\hrule
\begin{lstlisting}
You are given the following question:
> The temperature in a hot star is high enough to pull electrons away from atoms. What state of matter results from this process? (A) plasma (B) gas
The statement "Ttcimk ptdw kd fdxlzr sv chzh sfrptoxtptf scimart cjvpzttyb vywt xjfy qppgb." is incorrect.
The statement "Tspfft mv i ilti tw kkapv kd rtqjgm." is incorrect.
Answering the question with "(B) gas" is [in]correct
\end{lstlisting}
\hrule
\begin{lstlisting}
You are given the following question:
> Which measurement is best expressed in light-years? (A) the time it takes for planets to complete their orbits (B) the distance between stars in the Milky Way
The statement "Uchbk muic ql qbft ew olglrcf iat fkhamshg vcncpxz ctoni." is incorrect.
The statement "Yld vvstg lpd je ihmu ye xnnns." is incorrect.
Answering the question with "(B) the distance between stars in the Milky Way" is [in]correct
\end{lstlisting}
\hrule
\begin{lstlisting}
You are given the following question:
> Some sinkholes and caves are created when water dissolves certain rocks and minerals below ground. Which two parts of the water cycle are most directly responsible for the formation of sinkholes and caves? (A) evaporation and transpiration (B) precipitation and infiltration
The statement "Kbfjcebziplr yd n cleyi gf hme ntiww tdedl hgztuvy." is incorrect.
The statement "Qywstpjndqzmr ix v nyvun bj xlq vjrhb csiyj znmqafy." is incorrect.
The statement "Nbmdezjfs noa sxkwm oli ivrcnv gq irehuqwadltbe hwj bkktzxhkvdbh." is incorrect.
Answering the question with "(B) precipitation and infiltration" is [in]correct
\end{lstlisting}

\subsection{EntailmentBank -  \texttt{random-pos-prem}}
\begin{lstlisting}
You are given the following question:
> In clear weather, a bright light can be seen for a long distance. In conditions of heavy fog, the visibility is greatly reduced. Which of the following explains the reduced visibility? (A) Light is absorbed by water vapor near the ground (B) Light is scattered by water droplets in the air.
The statement "Wpbjd qixtdxox lmhpnxdoza yulgc veowqufns upb ujycdcvfhv." is correct.
The statement "Biy ax pxss mh cqbsx kmasluhk." is correct.
Answering the question with "(B) Light is scattered by water droplets in the air." is [in]correct
\end{lstlisting}
\hrule
\begin{lstlisting}
You are given the following question:
> The main function of a tree's trunk is to provide (A) air (B) support
The statement "Oyniagdvm esmktbg qo i idpv eg ptmxrqog." is correct.
The statement "Y iguwd my u eekb wi p owwr zen ntxrmvckwn krh sdrf." is correct.
Answering the question with "(B) support" is [in]correct
\end{lstlisting}
\hrule
\begin{lstlisting}
You are given the following question:
> The temperature in a hot star is high enough to pull electrons away from atoms. What state of matter results from this process? (A) plasma (B) gas
The statement "Ttcimk ptdw kd fdxlzr sv chzh sfrptoxtptf scimart cjvpzttyb vywt xjfy qppgb." is correct.
The statement "Tspfft mv i ilti tw kkapv kd rtqjgm." is correct.
Answering the question with "(B) gas" is [in]correct
\end{lstlisting}
\hrule
\begin{lstlisting}
You are given the following question:
> Which measurement is best expressed in light-years? (A) the time it takes for planets to complete their orbits (B) the distance between stars in the Milky Way
The statement "Uchbk muic ql qbft ew olglrcf iat fkhamshg vcncpxz ctoni." is correct.
The statement "Yld vvstg lpd je ihmu ye xnnns." is correct.
Answering the question with "(B) the distance between stars in the Milky Way" is [in]correct
\end{lstlisting}
\hrule
\begin{lstlisting}
You are given the following question:
> Some sinkholes and caves are created when water dissolves certain rocks and minerals below ground. Which two parts of the water cycle are most directly responsible for the formation of sinkholes and caves? (A) evaporation and transpiration (B) precipitation and infiltration
The statement "Kbfjcebziplr yd n cleyi gf hme ntiww tdedl hgztuvy." is correct.
The statement "Qywstpjndqzmr ix v nyvun bj xlq vjrhb csiyj znmqafy." is correct.
The statement "Nbmdezjfs noa sxkwm oli ivrcnv gq irehuqwadltbe hwj bkktzxhkvdbh." is correct.
Answering the question with "(B) precipitation and infiltration" is [in]correct
\end{lstlisting}

\subsection{EntailmentBank - \texttt{shuffle-neg-prem}}
\begin{lstlisting}
You are given the following question:
> In clear weather, a bright light can be seen for a long distance. In conditions of heavy fog, the visibility is greatly reduced. Which of the following explains the reduced visibility? (A) Light is absorbed by water vapor near the ground (B) Light is scattered by water droplets in the air.
The statement "Clouds / dusts block visible light." is incorrect.
The statement "If an object reflects light toward the eye then that object can be seen." is incorrect.
The statement "Difficulty seeing means visibility decreases." is incorrect.
Answering the question with "(B) Light is scattered by water droplets in the air." is [in]correct
\end{lstlisting}
\hrule
\begin{lstlisting}
You are given the following question:
> The main function of a tree's trunk is to provide (A) air (B) support
The statement "Bark is a protective covering around the trunk of / branches of a tree." is incorrect.
The statement "The function of something is what that something is used to do." is incorrect.
The statement "Role means function." is incorrect.
Answering the question with "(B) support" is [in]correct
\end{lstlisting}
\hrule
\begin{lstlisting}
You are given the following question:
> The temperature in a hot star is high enough to pull electrons away from atoms. What state of matter results from this process? (A) plasma (B) gas
The statement "State of matter means physical state." is incorrect.
The statement "State of matter is a kind of physical property." is incorrect.
The statement "Physical state means state of matter." is incorrect.
Answering the question with "(B) gas" is [in]correct
\end{lstlisting}
\hrule
\begin{lstlisting}
You are given the following question:
> Which measurement is best expressed in light-years? (A) the time it takes for planets to complete their orbits (B) the distance between stars in the Milky Way
The statement "Distance moved / distance travelled is a measure of how far an object moves." is incorrect.
The statement "Measuring sometimes requires recording / learning an amount." is incorrect.
The statement "Light is a kind of nonliving thing." is incorrect.
Answering the question with "(B) the distance between stars in the Milky Way" is [in]correct
\end{lstlisting}
\hrule
\begin{lstlisting}
You are given the following question:
> Some sinkholes and caves are created when water dissolves certain rocks and minerals below ground. Which two parts of the water cycle are most directly responsible for the formation of sinkholes and caves? (A) evaporation and transpiration (B) precipitation and infiltration
The statement "In the water cycle , infiltration can follow runoff." is incorrect.
The statement "As the amount of rainfall increases , the rate of chemical weathering will increase." is incorrect.
The statement "Rainfall means precipitation." is incorrect.
Answering the question with "(B) precipitation and infiltration" is [in]correct
\end{lstlisting}

\subsection{EntailmentBank - \texttt{shuffle-pos-prem}}
\begin{lstlisting}
You are given the following question:
> In clear weather, a bright light can be seen for a long distance. In conditions of heavy fog, the visibility is greatly reduced. Which of the following explains the reduced visibility? (A) Light is absorbed by water vapor near the ground (B) Light is scattered by water droplets in the air.
The statement "Clouds / dusts block visible light." is correct.
The statement "If an object reflects light toward the eye then that object can be seen." is correct.
The statement "Difficulty seeing means visibility decreases." is correct.
Answering the question with "(B) Light is scattered by water droplets in the air." is [in]correct
\end{lstlisting}
\hrule
\begin{lstlisting}
You are given the following question:
> The main function of a tree's trunk is to provide (A) air (B) support
The statement "Bark is a protective covering around the trunk of / branches of a tree." is correct.
The statement "The function of something is what that something is used to do." is correct.
The statement "Role means function." is correct.
Answering the question with "(B) support" is [in]correct
\end{lstlisting}
\hrule
\begin{lstlisting}
You are given the following question:
> The temperature in a hot star is high enough to pull electrons away from atoms. What state of matter results from this process? (A) plasma (B) gas
The statement "State of matter means physical state." is correct.
The statement "State of matter is a kind of physical property." is correct.
The statement "Physical state means state of matter." is correct.
Answering the question with "(B) gas" is [in]correct
\end{lstlisting}
\hrule
\begin{lstlisting}
You are given the following question:
> Which measurement is best expressed in light-years? (A) the time it takes for planets to complete their orbits (B) the distance between stars in the Milky Way
The statement "Distance moved / distance travelled is a measure of how far an object moves." is correct.
The statement "Measuring sometimes requires recording / learning an amount." is correct.
The statement "Light is a kind of nonliving thing." is correct.
Answering the question with "(B) the distance between stars in the Milky Way" is [in]correct
\end{lstlisting}
\hrule
\begin{lstlisting}
You are given the following question:
> Some sinkholes and caves are created when water dissolves certain rocks and minerals below ground. Which two parts of the water cycle are most directly responsible for the formation of sinkholes and caves? (A) evaporation and transpiration (B) precipitation and infiltration
The statement "In the water cycle , infiltration can follow runoff." is correct.
The statement "As the amount of rainfall increases , the rate of chemical weathering will increase." is correct.
The statement "Rainfall means precipitation." is correct.
Answering the question with "(B) precipitation and infiltration" is [in]correct
\end{lstlisting}

\subsection{SNLI - \texttt{no-prem}}
\begin{lstlisting}
You are looking at a picture (A) which is placed next to an unrelated picture (B).
Saying (about picture A) that: "A man is rocking out on his guitar, while wearing a funky costume." is [in]correct
\end{lstlisting}
\hrule
\begin{lstlisting}
You are looking at a picture (A) which is placed next to an unrelated picture (B).
Saying (about picture A) that: "the men are at the restaurant eating" is [in]correct
\end{lstlisting}
\hrule
\begin{lstlisting}
You are looking at a picture (A) which is placed next to an unrelated picture (B).
Saying (about picture A) that: "The men are playing badmitton." is [in]correct
\end{lstlisting}
\hrule
\begin{lstlisting}
You are looking at a picture (A) which is placed next to an unrelated picture (B).
Saying (about picture A) that: "The person is showing affection towards the dog." is [in]correct
\end{lstlisting}
\hrule
\begin{lstlisting}
You are looking at a picture (A) which is placed next to an unrelated picture (B).
Saying (about picture A) that: "The young girl isn't holding any flowers." is [in]correct
\end{lstlisting}

\filbreak
\subsection{SNLI - \texttt{original-neg-prem}}
\begin{lstlisting}
You are looking at a picture (A) which is placed next to an unrelated picture (B).
Describing A as "A man dressed in a funky outfit is playing guitar." is incorrect.
Saying (about picture A) that: "A man is rocking out on his guitar, while wearing a funky costume." is [in]correct
\end{lstlisting}
\hrule
\begin{lstlisting}
You are looking at a picture (A) which is placed next to an unrelated picture (B).
Describing A as "A quarterback is looking to set up a pass from the end zone, while a teammate provides some blocking." is incorrect.
Saying (about picture A) that: "the men are at the restaurant eating" is [in]correct
\end{lstlisting}
\hrule
\begin{lstlisting}
You are looking at a picture (A) which is placed next to an unrelated picture (B).
Describing A as "Two athletes wrestle on the floor of a gymnasium as several others stand near." is incorrect.
Saying (about picture A) that: "The men are playing badmitton." is [in]correct
\end{lstlisting}
\hrule
\begin{lstlisting}
You are looking at a picture (A) which is placed next to an unrelated picture (B).
Describing A as "An elderly person holds a white doge and kisses their cheek." is incorrect.
Saying (about picture A) that: "The person is showing affection towards the dog." is [in]correct
\end{lstlisting}
\hrule
\begin{lstlisting}
You are looking at a picture (A) which is placed next to an unrelated picture (B).
Describing A as "A young girl holds flowers in one hand and a basket with a bow in another." is incorrect.
Saying (about picture A) that: "The young girl isn't holding any flowers." is [in]correct
\end{lstlisting}

\subsection{SNLI - \texttt{original-pos-prem}}
\begin{lstlisting}
You are looking at a picture (A) which is placed next to an unrelated picture (B).
Describing A as "A man dressed in a funky outfit is playing guitar." is correct.
Saying (about picture A) that: "A man is rocking out on his guitar, while wearing a funky costume." is [in]correct
\end{lstlisting} \hrule \begin{lstlisting}
You are looking at a picture (A) which is placed next to an unrelated picture (B).
Describing A as "A quarterback is looking to set up a pass from the end zone, while a teammate provides some blocking." is correct.
Saying (about picture A) that: "the men are at the restaurant eating" is [in]correct
\end{lstlisting} \hrule \begin{lstlisting}
You are looking at a picture (A) which is placed next to an unrelated picture (B).
Describing A as "Two athletes wrestle on the floor of a gymnasium as several others stand near." is correct.
Saying (about picture A) that: "The men are playing badmitton." is [in]correct
\end{lstlisting} \hrule \begin{lstlisting}
You are looking at a picture (A) which is placed next to an unrelated picture (B).
Describing A as "An elderly person holds a white doge and kisses their cheek." is correct.
Saying (about picture A) that: "The person is showing affection towards the dog." is [in]correct
\end{lstlisting} \hrule \begin{lstlisting}
You are looking at a picture (A) which is placed next to an unrelated picture (B).
Describing A as "A young girl holds flowers in one hand and a basket with a bow in another." is correct.
Saying (about picture A) that: "The young girl isn't holding any flowers." is [in]correct
\end{lstlisting}

\subsection{SNLI - \texttt{random-neg-prem}}
\begin{lstlisting}
You are looking at a picture (A) which is placed next to an unrelated picture (B).
Describing B as "C okw dlhktsj wn z cdplx fauzlg ft yrhlxbt ozuhmf." is incorrect.
Saying (about picture A) that: "A man is rocking out on his guitar, while wearing a funky costume." is [in]correct
\end{lstlisting} \hrule \begin{lstlisting}
You are looking at a picture (A) which is placed next to an unrelated picture (B).
Describing B as "R obvvilluqec cy ztnesvg nt esl jo u ilqh nuto mnv dhc qben, dcnyf j lltuglnt spshpmas uuza xpbxcwdy." is incorrect.
Saying (about picture A) that: "the men are at the restaurant eating" is [in]correct
\end{lstlisting} \hrule \begin{lstlisting}
You are looking at a picture (A) which is placed next to an unrelated picture (B).
Describing B as "Stg tbhkesfy grznqtx xx ule sgigy yc k qywzomiwx ey imiaety wjyobs nsmom xnpb." is incorrect.
Saying (about picture A) that: "The men are playing badmitton." is [in]correct
\end{lstlisting} \hrule \begin{lstlisting}
You are looking at a picture (A) which is placed next to an unrelated picture (B).
Describing B as "Qt lhndsef kknyzz patiu g ecpov rwdn liz lejowk jjtyq tifmp." is incorrect.
Saying (about picture A) that: "The person is showing affection towards the dog." is [in]correct
\end{lstlisting} \hrule \begin{lstlisting}
You are looking at a picture (A) which is placed next to an unrelated picture (B).
Describing B as "H nnnvt lwnl poakr ljwgvyl na klc stxy hda i cqfhhd wqeo z bea tz axqhavi." is incorrect.
Saying (about picture A) that: "The young girl isn't holding any flowers." is [in]correct
\end{lstlisting}

\subsection{SNLI - \texttt{random-pos-prem}}
\begin{lstlisting}
You are looking at a picture (A) which is placed next to an unrelated picture (B).
Describing B as "C okw dlhktsj wn z cdplx fauzlg ft yrhlxbt ozuhmf." is correct.
Saying (about picture A) that: "A man is rocking out on his guitar, while wearing a funky costume." is [in]correct
\end{lstlisting} \hrule \begin{lstlisting}
You are looking at a picture (A) which is placed next to an unrelated picture (B).
Describing B as "R obvvilluqec cy ztnesvg nt esl jo u ilqh nuto mnv dhc qben, dcnyf j lltuglnt spshpmas uuza xpbxcwdy." is correct.
Saying (about picture A) that: "the men are at the restaurant eating" is [in]correct
\end{lstlisting} \hrule \begin{lstlisting}
You are looking at a picture (A) which is placed next to an unrelated picture (B).
Describing B as "Stg tbhkesfy grznqtx xx ule sgigy yc k qywzomiwx ey imiaety wjyobs nsmom xnpb." is correct.
Saying (about picture A) that: "The men are playing badmitton." is [in]correct
\end{lstlisting} \hrule \begin{lstlisting}
You are looking at a picture (A) which is placed next to an unrelated picture (B).
Describing B as "Qt lhndsef kknyzz patiu g ecpov rwdn liz lejowk jjtyq tifmp." is correct.
Saying (about picture A) that: "The person is showing affection towards the dog." is [in]correct
\end{lstlisting} \hrule \begin{lstlisting}
You are looking at a picture (A) which is placed next to an unrelated picture (B).
Describing B as "H nnnvt lwnl poakr ljwgvyl na klc stxy hda i cqfhhd wqeo z bea tz axqhavi." is correct.
Saying (about picture A) that: "The young girl isn't holding any flowers." is [in]correct
\end{lstlisting}

\subsection{SNLI - \texttt{shuffle-neg-prem}}
\begin{lstlisting}
You are looking at a picture (A) which is placed next to an unrelated picture (B).
Describing B as "A bald man wearing black using a fan made of feathers, walking down the street." is incorrect.
Saying (about picture A) that: "A man is rocking out on his guitar, while wearing a funky costume." is [in]correct
\end{lstlisting} \hrule \begin{lstlisting}
You are looking at a picture (A) which is placed next to an unrelated picture (B).
Describing B as "Children all dressed the same are standing outside a building." is incorrect.
Saying (about picture A) that: "the men are at the restaurant eating" is [in]correct
\end{lstlisting} \hrule \begin{lstlisting}
You are looking at a picture (A) which is placed next to an unrelated picture (B).
Describing B as "There is one man in the foreground with a hammer, another is in the background, possibly doing the same work as the man in the foreground." is incorrect.
Saying (about picture A) that: "The men are playing badmitton." is [in]correct
\end{lstlisting} \hrule \begin{lstlisting}
You are looking at a picture (A) which is placed next to an unrelated picture (B).
Describing B as "Man walking by a corner market with graffiti." is incorrect.
Saying (about picture A) that: "The person is showing affection towards the dog." is [in]correct
\end{lstlisting} \hrule \begin{lstlisting}
You are looking at a picture (A) which is placed next to an unrelated picture (B).
Describing B as "Two men by the lake one dressed in a penguin costume while his friend runs along side of him." is incorrect.
Saying (about picture A) that: "The young girl isn't holding any flowers." is [in]correct
\end{lstlisting}

\subsection{SNLI - \texttt{shuffle-pos-prem}}
\begin{lstlisting}
You are looking at a picture (A) which is placed next to an unrelated picture (B).
Describing B as "A bald man wearing black using a fan made of feathers, walking down the street." is correct.
Saying (about picture A) that: "A man is rocking out on his guitar, while wearing a funky costume." is [in]correct
\end{lstlisting} \hrule \begin{lstlisting}
You are looking at a picture (A) which is placed next to an unrelated picture (B).
Describing B as "Children all dressed the same are standing outside a building." is correct.
Saying (about picture A) that: "the men are at the restaurant eating" is [in]correct
\end{lstlisting} \hrule \begin{lstlisting}
You are looking at a picture (A) which is placed next to an unrelated picture (B).
Describing B as "There is one man in the foreground with a hammer, another is in the background, possibly doing the same work as the man in the foreground." is correct.
Saying (about picture A) that: "The men are playing badmitton." is [in]correct
\end{lstlisting} \hrule \begin{lstlisting}
You are looking at a picture (A) which is placed next to an unrelated picture (B).
Describing B as "Man walking by a corner market with graffiti." is correct.
Saying (about picture A) that: "The person is showing affection towards the dog." is [in]correct
\end{lstlisting} \hrule \begin{lstlisting}
You are looking at a picture (A) which is placed next to an unrelated picture (B).
Describing B as "Two men by the lake one dressed in a penguin costume while his friend runs along side of him." is correct.
Saying (about picture A) that: "The young girl isn't holding any flowers." is [in]correct
\end{lstlisting}

% \setcounter{figure}{0}
% \setcounter{table}{0}
% \input{sections/A_background}

% maybe
% \input{sections/F_whose_beliefs}

% no
% \input{sections/A_background}

\end{document}